\documentclass[manuscript,screen,preprint,nonacm]{acmart} 

\usepackage{graphicx} % Required for including images
\usepackage{subcaption} % Subfigures 
\usepackage{multicol}
\usepackage{multirow}

\newcommand{\indep}{\perp \!\!\! \perp} 
\AtBeginDocument{%
  \providecommand\BibTeX{{%
    \normalfont B\kern-0.5em{\scshape i\kern-0.25em b}\kern-0.8em\TeX}}}

\setcopyright{acmcopyright}
\copyrightyear{2022}
\acmYear{2022}
\acmDOI{}

\acmJournal{CSUR}

\begin{document}

\title{Data Representativity for Machine Learning and AI Systems}

\author{Line H. Clemmensen}
\authornote{Equal contributions}
\email{lkhc@dtu.dk}
\orcid{0000-0001-5527-5798}
%\email{lkhc@dtu.dk}
\affiliation{%
  \institution{Technical University of Denmark}
  \streetaddress{Richard Petersens Plads 324}
  \city{Kgs. Lyngby}
  \country{Denmark}
  \postcode{2800}
}

\author{Rune D. Kjærsgaard}
\authornotemark[1]
\affiliation{%
  \institution{Technical University of Denmark}
  \streetaddress{Richard Petersens Plads 324}
  \city{Kgs. Lyngby}
  \country{Denmark}
  \postcode{2800}
  }
\email{rdokj@dtu.dk}

\renewcommand{\shortauthors}{Clemmensen and Kjærsgaard}

%% article

\begin{abstract}
Data representativity is crucial when drawing inference from data through machine learning models. Scholars have increased focus on unraveling the bias and fairness in models, also in relation to inherent biases in the input data. However, limited work exists on the representativity of samples (datasets) for appropriate inference in AI systems. This paper reviews definitions and notions of a \emph{representative sample} and surveys their use in scientific AI literature. We introduce three measurable concepts to help focus the notions and evaluate different data samples. Furthermore, we demonstrate that the contrast between a representative sample in the sense of coverage of the input space, versus a representative sample mimicking the distribution of the target population is of particular relevance when building AI systems. Through empirical demonstrations on US Census data, we evaluate the opposing inherent qualities of these concepts. Finally, we propose a framework of questions for creating and documenting data with data representativity in mind, as an addition to existing dataset documentation templates.
\end{abstract}



%\ccsdesc[500]{Theory of computation~Data structures design and analysis}

%\ccsdesc[500]{Computing methodologies}
%\ccsdesc[100]{Artificial intelligence}

%\ccsdesc[500]{Computer systems organization~Embedded systems}


\maketitle

\section{Introduction}
Machine learning and AI systems are increasingly governing important decisions affecting individuals at all levels of society. These automated decision frameworks have demonstrated various unwanted consequences as a result of biased data \cite{phillips2011otherrace, lum2016predict, Buolamwini, Raji, FacebookAd, Lowry, Obermeyer}. Oftentimes these systems are trained on samples (datasets) from a larger population. Biased results can arise if the sample does not accurately represent the target population, or if there is a lack of sufficient representation for subgroups within the data. While the literature of data bias in machine Learning and artificial intelligence (AI) systems is rich \cite{suresh2019framework}, there exists only limited work on the connections between \emph{data representativity} and AI systems. Terms like \emph{representative sample} are used ubiquitously in the literature, often without further specification on the details or effects of this representativity. This paper analyzes and surveys data representativity in scientific literature relating to machine learning and AI systems by investigating how different notions of representativity are used and what effects adhering to different notions of data representativity has in relation to appropriate inference.

The term \emph{representative sample} is an overloaded term and a generally accepted definition of what constitutes a \emph{representative sample} (subset of observations) is hard to find in the literature. A few examples demonstrate that at least a couple of definitions of \emph{representative sample} exist. The most general definition we found is from D'Excelle (2014) and states \emph{"“Representative sampling” is a type of statistical sampling that allows us to use data from a sample to make conclusions that are representative for the population from which the sample is taken."} \cite{D'Excelle}. However, this definition leaves us with the important question of what we mean by representative. The following two examples of definitions clarify this point. 1) Meriam Websters' online dictionary says: \emph{"Sampling in which the relative sizes of sub-population samples are chosen equal to the relative sizes of the sub-populations."} \cite{MeriamWebster}. 2) An online portal disseminating elementary statistics to graduate students writes \emph{"A representative sample is where your sample matches some characteristic of your population, usually the characteristic you’re targeting with your research."} \cite{StatisticsHowto}. These examples illustrate that as we unfold the meaning of representative, questions arise, like what the target population is and which attributes/characteristics/sub-populations are relevant as well as how to measure a match between a sample and a population. OECD (Economic Co-operation and Development)'s definition of a representative sample acknowledges that several notions exist: \emph{"In the widest sense, a sample which is representative of a population. Some confusion arises according to whether “representative” is regarded as meaning “selected by some process which gives all samples an equal chance of appearing to represent the population”; or, alternatively, whether it means “typical in respect of certain characteristics, however chosen"."} \cite{OECD}. Some of these ambiguities are linked: The definition of representative is linked to the target of the system/research/analysis and dictates which attributes, sub populations, and representative measures are of relevance. In this paper, we review various interpretations and notions of the term \emph{representative sample} and link these to mathematical measures. Subsequently, we measure the match between the notions and the target of the analysis by looking at performance, diversity, and fairness metrics.

In 1979-1980 Kruskal and Mosteller wrote four papers on the term \emph{representative sampling} with the motivation to unravel its ambiguities and imprecision \cite{KruskalMosteller1979-1, KruskalMosteller1979-2, KruskalMosteller1979-3, KruskalMosteller1979-4}. In addition, they called for caution as well as more specific expressions when referring to a representative sample. As they noted: "The reason for so much effort on one term is that the idea of representativeness is closely related to basic notions of statistical inference". In this paper, we take a closer look at data representativity for recent machine learning and artificial intelligence (AI) systems and before advancing, we will dwell on the nature of studies in AI and what this means for inference. AI systems are built both on observational data and on data from experiments gathered with the purpose of training the AI. Whereas randomized controlled experiments/trials are truly random samples, observational studies need to be carefully designed to tackle their inherent haphazardness \citep{Rosenbaum2010}. In observational studies, matching is performed to make treatment and control groups comparable, but unlike for experimentation, there is no basis for assuming that this extends to unmeasured factors \citep{Rosenbaum2010, Montgomery219}. Experimental studies are often used to make causal inferences, a basis which dates back to R.A. Fisher (1935) \citep{Fisher1935}. However, causal relations can also be established through observational studies, like for example the link between smoking and lung cancer \citep{Cornfield2009}. We will therefore not further distinguish between the nature of the data or the AI systems.

As we draw conclusions from data or make predictions in artificial intelligence (AI) systems trained on data, it is important to understand what these data represent, and which inferences we can make. AI systems or machine learning (ML) models for decision making are widely used in industry and research, but care is not always put to the origin of the data, on which the systems are trained. This is for example seen in big data, where more data are considered better, and data often originate from a historical collection performed for e.g., control purposes or from scraping available internet sources rather than having been collected for the purpose, which it is later used for \citep{Boyd2011, Kulahci2020, Huang2021, Beresewicz2017}. Other examples are more general for ML/AI and include representation bias stemming from the way we define and sample from a population, evaluation bias stemming from benchmark datasets with inherent biases, population bias when the distribution of attributes differ between dataset and target population, and sampling bias stemming from non-random sampling of subgroups \citep{Mehrabi2019, Olteanu2019, Suresh2019}.

Amongst other, Kruskal and Mosteller found that \emph{representative sample} was used as an assertive to underline a point without any scientific reasoning. Historically, the ImageNet competition has had a kind of implicit assertive, where scientists believed good results on the ImageNet dataset would mean good results for other image recognition tasks as well \citep{Mehrabi2019, DotanMilli2020}. Torralba and Efros empirically illustrated in their paper 'Unbiased Look at Dataset Bias' (2011) that generalizations supporting this assertive were not necessarily a given, and described their findings as "if we add training data that does not match the biases of the test data this will result in a less effective classifier" \citep{Torralba2011}. 

Recently, focus has been put on the lack of transparency around dataset design and collection procedures as well as efforts to unbias existing datasets like e.g., the ImageNet \citep{Mehrabi2019, Yang2020}. We will investigate these initiatives as well as the notions of a representative sample within the AI community. We have found sampling theories from the disciplines of analysis of physical material, design of experiments, as well as surveys in social sciences useful in terms of analyzing current practices and relating these to the ongoing work within ML and AI, where the historical emphasis on data representativity has been smaller.

To summarize, our contributions in this paper are:
 \begin{itemize}
\item We provide an overview of the interdisciplinary topic of data representativity, organise the various notions of representativity, link mathematical measures to the notions when possible, and propose the use of three measurable concepts.
\item We describe and discuss relevant notions of representativity in literature and review their use in papers introducing datasets from the NeurIPS 2021 Track on Datasets and Benchmarks and ICCV 2021.
\item We demonstrate contrasting perspectives on data representativity by empirically comparing two measurable concepts with opposing notions of representativity.
\item We propose a framework of questions for creating and documenting data with representativity in mind.
\item We provide new research directions on data representativity in ML and AI.
\end{itemize}

The rest of the paper is organized as follows. First, through literature about representative sampling, we will outline the general notions of a 'representative sample' (Section \ref{sec:notions}), give examples of their use in recent ML and AI literature, add mathematical measures for each notion, when possible, and propose to use three measurable concepts inn their place. In Section \ref{sec:Survey} we review the notions of representative sample used in the papers from the datasets and benchmarks track at NeurIPS 2021 and new benchmark datasets from ICCV 2021. Throughout these investigations, we find opposing opinions of sampling for coverage of the input space vs. probability sampling mimicking population distributions, which correspond to two of the measurable concepts. Consequently, we make empirical investigations demonstrating the qualities these opposing notions/concepts hold in Section \ref{sec:data}. Finally, we suggest a framework for addressing data representativity in datasheets in Section \ref{sec:framework} and round off with a discussion in Section \ref{sec:discussion}.

\section{Notions of a 'representative sample'}\label{sec:notions}
Since there is no specific, mathematical definition of a representative sample, and initial investigations identified at least a couple of different notions of what a representative sample is, we will review differing notions here. 

Kruskal and Mosteller identified six notions/usages of a 'representative sample' in their first surveys from 1979 \citep{KruskalMosteller1979-1, KruskalMosteller1979-2}: An assertive acclaim, absence of selective forces, a miniature of the population, an observation 'typical' or 'ideal' of the (sub)population, coverage of a population by the sample, and a reference to a sampling method later on specified in details. The sixth is a special notion in scientific writing, whereas the first five were found in both non-scientific as well as scientific writing. We will use this framing here, and link more recent literature to these and add examples of their use in literature. We add existing mathematical or formal definitions belonging to the notions in the subsequent section. 

Finally, we also add two novel notions we found in AI literature. We call them the copycat and no notion. Copycat refers to the creation of synthetic data representative of a target population. No notion refers to vague or no mentioning of representativity and likewise also no mentioning of non-representativity or limitations of the data representativity. The latter may seem harmless when presenting new datasets, but as the data is re-used, this can become harmful and an implicit notion of an assertive claim can grow in its place.

\subsection{The assertive claim (the Emperor's new clothes)} The assertive claim, as described in the introduction, is used as an assertive to underline a point without any scientific reasoning and is dangerous both as a conscious acclaim and a subconscious notion when it comes without specification. It is recommended to avoid unjustified and unspecified use. We mentioned ImageNet as a historical example of an assertive claim of a representative sample \citep{Mehrabi2019, DotanMilli2020}. Despite the broad acknowledgment of the ImageNet case as a cautionary tale on data representativity, the assertive notion continues to appear even in recent literature from acknowledged publication venues. One example is from the datasheet of a publication from the NeurIPS 2021 Track on Datasets and Benchmarks regarding time-sensitive questions \cite{survey_10}: "It’s sampled from large Wikipedia passages, it’s representative of all the possible temporal-sensitive information."

\subsection{The miniature (the model train set)} 
The miniature is best captured by Meriam-Webster's definition: "the relative sizes of sub-population samples are chosen equal to the relative sizes of the sub-populations" \cite{MeriamWebster}. This is a sample of the target population perfectly mimicking every (relevant) aspect (characteristic/distribution) of the population.

The miniature population has strong ties to the theory of sampling of physical material also related to chemical or biological analysis \citep{Gy1998, Petersen2005}. One of the guiding principles in the theory of sampling is to have as homogeneous a population (lot) as possible in order for a sample anywhere in the lot to mimic the lot best possible, which in turn minimizes sampling errors. 

In other fields, it is common to subdivide the space into smaller groups, until each group exhibits homogeneity, and then randomly sample a miniature or a sample representative of that group with probability equal to the proportion of that group in the population \cite{bornstein2013sampling}. This is also referred to as strata sampling (simple random sampling within mutually exclusive groups of the target population/strata) or cluster sampling (random sampling of clusters/strata in the population and inclusion of all samples for the selected clusters) in fields like survey analysis \citep{Gideon2012-5}. Ghojogh et al show that strata sampling always has lower variance than that of simple random sampling, in particular when strata have very different characteristics \citep{ghojogh2020}. However, defining homogeneity in terms of subgroups may be delicate and constructing meaningful groups/strata is difficult if the population values/distributions are unknown. 

Sampling from distributions is another way to construct miniature samples \citep{shaw2006, MacKay2003}. Sampling from distributions, and not least joint distributions, gives the possibility of matching distributions between sample and population rather than matching simpler characteristics, like e.g. averages. In high dimensions, these methods do suffer computationally, however. As a non-parametric alternative it is possible to sample from densities \cite{kjaersgaard2021sampling,ros2016dendis}.

It is also possible to make a sample mimic certain characteristics of the population by sampling enough random samples to obtain a convergence in the measure of interest \citep{Blatchford2021}.

Recently, Yang et al (2020) \citep{Yang2020} proposed a framework to balance the demographics of ImageNet, but they also stated that this is only possible for one attribute at a time, as sub-categories will have too few samples if balancing across multiple attributes (e.g. race and gender). In consequence, the miniature analogy in itself breaks down, as we cannot account for all factors in the miniature, in particular not as the miniature decrease in sample size.

A concrete use of the miniature notion is seen in \citep{Machado2020} where Machado et al predict suicide attempts in what they refer to as a representative sample of the US population. They write: "a representative sample of the adult population of the United States, oversampling black people, Hispanic individuals, and young adults aged 18– 24 years. ... Weighted data were adjusted to be representative of the civilian population ... data were weighted to reflect design characteristics of the NESARC and account for oversampling." The miniature notion is apparent in terms of reweighing characteristics to match the distribution of the population of interest. A certain notion of coverage and absence of selective forces can also be seen in terms of age and race, for which specific sampling strategies (oversampling) have been taken. This example illustrates that several notions are often used together, something we also note in our survey in Section \ref{sec:Survey}.

In 'Understanding the Demographics of Twitter Users' by Mislove et al (2011) they conclude that Twitter users are not representative of the US population based on argumentation of non-matching demographic distributions for geography, gender, and race/ethnicity \citep{Mislove2011}. This notion is related to that of a miniature, and we note that a dismissal of the representativity is in essence easier than proving it holds. However, even a dismissal of a sample as representative is limited to our understanding of the population. An understanding which for example is limited as explained by Taleb's Black Swan theory \citep{Taleb2007, Taleb2020} about human's rationalization of rare and unpredictable events. Ruths and Pfeffer later on proposed eight steps to reduce biases and flaws in social media data \citep{Ruths2014}, parts of these relate to the data collection and its documentation (similar to datasheets for datasets \citep{Gebru2021}), and another part relates to correction for biases by population matching (miniature notion) or robustness testing across time and different samples.

\subsection{Absence or presence of selective forces (justice balancing the scales)} Absence or presence of selective forces means that the sample is random as no forces are in play to select or de-select any specific types of observations in the target population; implying the purpose is to make inference about the target population, not the sample. 

This notion ties to experimental modeling and coverage as follows. In the design of experiments literature, controllable factors and uncontrollable factors are distinguished \citep{Montgomery219}. The controllable ones are indeed controlled to design as small an experiment as possible, yet with a suitable amount of observations and an appropriate coverage of the input space in order to make inference and optimize the response/output as a function of the controllable factors. Too many controlled factors make it hard to access all cross populations, and in addition there is no way of exhausting all possibilities. 

Selective factors can also be uncontrollable or in the worst case go unnoticed. Examples of these are time-drifts in a production or non-response in surveys. These can pose problems to the statistical inference drawn from data. If observable, we can manage through our sampling design or sometimes even through post processing of data. However, unobserved or even unnoticed factors impose serious risks of bias and confounding. 

In surveys, non-response is considered a substantial source of error caused by selection, one that is not directly related with the sampling. Selective forces can also influence survey responders through e.g., an interviewer effect. Errors stemming from such selective forces can lead to potential biases, and several corrective efforts are usually applied to adjust for these \citep{Gideon2012}. 

Selective sampling can also be performed on purpose, in survey sampling such examples are: quota sampling, purposive sampling, and referral sampling. These sampling designs are non-random and generalizations are therefore challenged, but sometimes samples of interest are so few, or participation recruitment so difficult, that convenience sampling designs can come in handy \citep{Gideon2012}.

In Kelly et al's 2019 opinion paper 'Key challenges for delivering clinical impact with artificial intelligence' \citep{Kelly2019}, they mention representative sample as follows: "The curation of independent local test sets by each healthcare provider could be used to fairly compare the performance of the various available algorithms in a representative sample of their population." This notion of a representative sample speaks to some absence of selective forces in that it is believed each healthcare provider is best off providing its own sample, representative for their population, thus arguing for local models specific for a geographic area with specific demographics. Furthermore, distribution shifts are mentioned as a challenge for the AI models in healthcare, not only across healthcare providers, but also across time. This methodological discussion of whether a population should be seen as fixed or whether it itself is taken from an underlying stochastic process has ties all the way back to discussions from the 1903 ISI Berlin meeting (World Statistical Congress) \citep{KruskalMosteller1979-4}.

\subsection{Typical/ideal (Superman/Superwoman or the average man/woman)} Typical/ideal refers to typical or ideal exemplars which represent a population or subgroups of a population. This is not necessarily in a statistical sense, but may mean close to the average. An example is that in \citep{Lee2010}, where cluster centers from Gaussian mixture models are sampled as representative observations of a larger dataset. In addition, a ML method like archetypal analysis \citep{AA1994} carries some of this notion: Archetypes in the data are identified as linear combinations of the observations which describes a convex hull off the observations.

Another example of the notion of typical observations is from NeurIPS 2021, where typical names are sampled for construction of a dataset: "For each race and gender, we chose the top ten first names based on their overall frequency and representation within each group, excluding unisex names and names that differed by only one character." \cite{Survey_43}. As the authors state: "The names we selected were derived using real-world data on demographic representations of first names, however demographic representation does not necessarily correlate with implicit stereotypical associations".

We also found a use of a representative sample, meaning a sample representative of a specific target. In online tracking, this is used to help overcome occlusions when following a target in a video \citep{Ou2018}. This meaning is most related to that of typical exemplars, here typical of a specific target of interest.

\subsection{Coverage (Noah's Ark)} Coverage seeks to include the heterogeneity of the population in the sample. A strong requirement for coverage would be that the sample should contain at least one observation from each relevant partition of the population. In contrast to the miniature, coverage does not require proportions within partitions to match those of the population. Harry V. Roberts suggested in 1971 sampling following the coverage notion in order to select a committee and avoid conscious and unconscious biases from appointing authorities \citep{Roberts1971}. For the committee purpose, there is certain overlap with the typical/ideal notion. Along these lines, coverage is more about producing \emph{representativeness} than about obtaining a likeness with the original population.

Density-based sampling approaches have proven useful under the coverage notion of representative sampling, where density estimates can be used to asses population imbalances and use this information for sampling to cover the heterogeneity of the population in the sample \cite{kjaersgaard2021sampling,ros2016dendis}, or to reduce noise and improve performance in imbalanced classification settings \cite{hou2019density}.

An example where we meet the notion of coverage is in one of the recent proposals to address the lack of transparency around dataset collection and design in ML/AI, namely in \emph{datasheets for datasets} by Gebru et al (2021) \citep{Gebru2021}. One of the questions they propose concerns data representativity, and says:
"Does the dataset contain all possible instances or is it a sample (not necessarily random) of instances from a larger set? If the dataset is a sample, then what is the larger set? Is the sample representative of the larger set (e.g., geographic coverage)? If so, please describe how this representativeness was validated/verified. If it is not representative of the larger set, please describe why not (e.g., to cover a more diverse range of instances, because instances were withheld or unavailable)." Apart from a clear notion of coverage, the description concerns some of the historical issues noted by the earliest endeavors of Anders Kiær (Director of Statistics Norway during 1877-1913) to go from full census to a representative sample, namely, how do we measure the representativeness? \citep{KruskalMosteller1979-4}. Coverage may or may not be what we go for, but if we go for it, how do we measure coverage, in particular considering joint distributions from several attributes? For example if mean values or min/max of each attribute match between sample and population, this does not imply that the distributions of each attribute match between sample and target population. This only becomes more complex if we consider the joint distributions of the attributes. Second, we should note that if we strictly go for coverage, then distributions between sample and population most likely do not match, and e.g., variance or mean estimates based on the sample will differ. On the other hand, coverage has an intuitive attraction when it comes to inclusion and equality. We will demonstrate these aspects empirically in Section \ref{sec:data}.

In a benchmark data publication with focus on real-world images \citep{Li2019}, we additionally see a notion of coverage: "objectives for underwater image collection: ... a diversity of underwater scenes, different characteristics of quality degradation, and a broad range of image content should be covered."

Sampling with a notion of coverage in mind often means combining non-random and random sampling methods, whereas sampling with a miniature in mind often means using random probability sampling, for example strata sampling.

Coverage is also usually constructed purposefully to not mimic the underlying population, but rather to include the heterogeneity in the population, and this is often a preferred notion when fairness is part of the purpose of the modeling. In literature, some of the closest mathematical measures of coverage are those of diversity \citep{celis2016fair}.

\subsection{Reference to sampling, later on specified} 
With this notion, the term 'representative sample' in itself becomes a 'vague term', and the exact meaning is specified in the context. Kruskal and Mosteller recommended this use of the term \emph{representative sample}, bearing in mind that it needs always a specification. In their mind, the specification refers to the method of sampling, i.e., a description of how the data have been obtained. Apart from the sampling method/procedure we recommend also specifying the original population, the purpose of the sampling, and herein the notion (later refined to measurable concept)  under which the sample is taken.

Another question Gebru et al propose to answer in a datasheet refers to the method of sampling \cite{Gebru2021}: "If the dataset is a sample from a larger set, what was the sampling strategy (e.g., deterministic, probabilistic with specific sampling probabilities)?" Underlining the historical recommendations for a specification of sampling method when referring to a 'representative sample'. We will add, that any dataset is a sample of a larger set or population. In fact, this question may also give some of the answers to the question of how the data is representative or not, as these answers heavily depend on the sampling strategy. Hopefully, answers are also well aligned with the first question in the motivation part of Gebru et al's datasheet, namely "For what purpose was the dataset created?" For some purposes, small sets of data, not generally representative of the entire population in question, can be good enough. Subsets of data may show that some characteristic thought to be absent or rare is in fact more frequent, or vice versa, that something thought of as universal is in fact missing to at least some degree. These subsets may be representative of only a part of the underlying population and thus form basis to dismiss one of the mentioned hypotheses, but not to draw any further inference about the entire population, see also \citep{KruskalMosteller1979-3} for examples. With open source datasets, we should be careful, as the purpose or the hypothesis means we have collected specific data to enlighten us, and this data may not be useful to draw inference for other hypotheses or purposes.

As the method of sampling is specified, the notion of representativeness should be made clear and the reproduceability of the data/study possible. However, it is the notion and the purpose of the study that makes way for mathematical measures of the representativeness.

\subsection{The copycat (synthetically generated)}
This notion is used when real-world data (or parts of it) are copied or mimicked through synthetic data generation methods. In these settings the synthetic data are often claimed to represent the real-world data for instance by matching distributions. Alternatively, the synthetic data can be used to specifically target underrepresented regions of the original population distribution and thus be claimed more representative of uncommon instances than the original real-world data. Thus, synthetic data generation frameworks allow great flexibility and provide excellent test-beds for the study of data representativity. A recent example can be found in a paper publishing a novel text dataset \cite{Survey_64}: "We took advantage of the synthesis pipeline to showcase how datasets can be constructed with properties that deliberately differ from real world distributions. Notably, we include samples of individuals with common (e.g., scientist) as well as uncommon occupations (e.g., spy)... and designed SynthBio to be more balanced with respect to gender and nationality compared to the original WikiBio dataset. ... Our paper takes the stance that in addition to evaluating on the world as it is, researchers benefit from having the option to evaluate their models on a more uniform distribution of the population. Synthesizing novel datasets is one technique that serves this goal. ... In addition, undesirable bias in real-world data, especially with respect to underrepresented groups, can be controlled in synthetic data, enabling evaluation of model performance on comparatively rare language phenomena". We note that there are also notions of both a miniature and coverage in this example.

\subsection{No notion}
This notion, or rather lack thereof, indicates that it is simply not mentioned how or what the data may be representative of, or that it has no notion by not mentioning the limitations of the dataset. A newer benchmark dataset (from ICCV 2019) gives us an example of no notion of representativity \citep{Uy2019}. They describe one of their contributions as: "A new object dataset from meshes of scanned real-world scene for training and testing point cloud classification". They indicate that the real-world scans of objects are more representative of problems expected to occur in vision tasks than computer generated object scans. This is undoubtedly true, but when it comes to the dataset as a benchmark of real-world scenes, the data representativity is more unclear. The real world examples consist of 15 categories of indoor objects; "we manually filter and select objects for 15 common categories". It is unclear how the categories were chosen or what they are representative of in terms of a larger population of common indoor objects. We recommend more explicit descriptions of data representativity, sampling and its purpose, see also \citep{Gebru2021}. In some circumstances outlining the limitations of the data representativity may be more sensible than outlining the target population.

\subsection{Measurable concepts}
The notions do not provide clear definitions and sometimes several notions share the same underlying concept. Additionally, mathematical measures are not necessarily applicable to all notions. In this section we relate the notions to overarching concepts and connect these to mathematical measures that can be used to assess the concepts. In addition, we provide mathematical measures which can be used to asses the impact of data representativity on fairness.

We define three operational concepts for data representativity, see Table \ref{tab:concepts}. 1) A sample as a \emph{reflection} of the target population - mimicking the population distribution. The representativity can be measured by comparing the distributions of sample and target or by comparing specific measures (like averages) of interest. 2) A sample providing \emph{coverage} of the population. The coverage of the sample can be measured through existing diversity measures of the sample (like geometric diversity or entropy). 3) Samples as \emph{representatives} of subgroups in the population, where the representativity e.g., can be measured through cluster metrics like the average distance to the representative within the subgroup.

The miniature and coverage notions naturally fit into the reflection and coverage concepts, respectively. Synthetic data (copycat) are often devised according to a reflection concept, but can in also be devised according to a coverage concept. The notion of selective forces likewise fits into either the reflection or coverage concept depending on the aim. The reference to sampling notion is also context dependent conditional on the specified sampling procedure. This notion can adhere to any of the three concepts depending on the specified sampling methodology and aim of the study. 

\begin{table}[h!]
\caption{Overview of notions, concepts and related mathematical measures.}
\label{tab:concepts}
\begin{tabular}{p{1.9cm}|p{2cm}|p{4.8cm}|p{4.8cm}}
\hline
\toprule
     \textbf{Concept} & \textbf{Notion} & \textbf{Description} & \textbf{Examples of existing mathematical measures}  \\
     \midrule
     - & Assertive claim & Claiming representativeness without justification &  None - Avoid\\ \midrule
     - & No notion & No indication of data representativity & None - Avoid \\ \midrule
     - & Reference to sampling & Special notion specified in context of sampling method & Context dependent \\ \midrule
     \multirow{3}{*}{Reflection} & Miniature & Sample mimics population distribution & Averages and average predictions as well as distributional comparisons between sample and population \\ \cline{2-3}
      & Selective forces & Truly random sample in observational studies like e.g. surveys &  \\ \cline{2-3}
      %& Reference to sampling & Description of sampling method & \\ \cline{2-3}
      & Copycat & Synthetic data created to mimic real-world data distribution & \\ \midrule
     \multirow{3}{*}{Coverage} & Coverage & Sample provides coverage by broadly representing the heterogeneity/diversity of the population & Diversity measures of the sample e.g. geometric coverage \\ \cline{2-3}
     & Selective forces & Truly random samples in experimental studies &  \\ \cline{2-3}
     %& Reference to sampling & Description of how the sample is obtained & \\ \cline{2-3}
     & Copycat & Synthetic data created for balanced coverage of real-world dataspace & \\\midrule
     
    \multirow{1}{*}{Representatives} & Typical/ideal & Single observations are representatives of a group in the population &  The representatives are e.g. approximated by the mean, median or mode of the group \\ %\cline{2-3}
      %& Selective forces & Truly random sample in observational studies like e.g. surveys &  \\ \cline{2-3}
      %& Reference to sampling & Description of sampling method & \\ \cline{2-3}
      %& Copycat & Synthetic data created by variation on archetypical patterns or observations & \\ 
     
  %   Representatives & Typical/ideal & Single observations are representatives of a group in the population & The representatives are e.g. approximated by the mean, median or mode of the group \\ 
     \bottomrule
\end{tabular}
\end{table}

The notion of a representative sample as an assertive claim, whether explicit or implicit should be avoided as it is not measurable. Not having a notion is likewise not recommended as it is not measurable and may lead to an implicit assertive use of the dataset. 
While Kruskal and Mosteller preferred to use a notion of a 'representative sample' as a vague term with the sampling procedure specified later, we recommend clearly stating a motivation for data representativity and to subsequently thoroughly document the sampling procedure and methods. We argue that a more explicit use of one of the measurable concepts of representativity will make the aim clearer; giving the sampling documentation a context in which it can be evaluated.  

The reflection and coverage concepts often work from contrasting perspectives on data representativity and carry different inherent advantages and disadvantages. We will demonstrate these in Section \ref{sec:data}.
 
\subsubsection{Reflection}
This concept may be assessed in various ways. As a first approach, statistical tests on averages and average predictions can give an indication of generalization between sample and population. Additional central tendency measures like median and mode and statistical dispersion measures like variance and interquartile range may also be analyzed. Furthermore, the notion may be examined by analyzing the distributions, for instance measuring the distributional departure of the sample from the population. This departure can be measured through the $\ell_{\infty}$ norm equivalent to the Kolmogorov-Smirnov (KS) statistic \cite{lehmann2005testing}. Most distributional measures operate in one dimension, but some can be extended to compare multivariate distributions. For instance the generalization of the KS two sample statistic for 2D and 3D distributions due to Peacock \cite{peacock1983two}. Other tests, like the maximum mean discrepancy (MMD) \cite{gretton2012kernelMMD} are designed for comparing multidimensional distributions. Generally, the tests for comparing multidimensional distributions are computationally expensive for large, high-dimensional samples. Another popular distributional distance measure is the general Wasserstein distance \cite{vaserstein1969markovEMD1,kantorovich1960mathematicalEMD2} given by:
\begin{equation}
    W_p(\mu,\nu) =\left(\underset{\mathrm{\pi} \in \Gamma(\mu,\nu)}{\mathrm{inf}} \int_{M \times  M} d(x,y)^p \mathrm{d\pi}(x,y) \right)^{1/p},
    \label{Eq:EMD}
\end{equation}
where $p\geq1$ and $W_p$ is the $p^\mathrm{th}$ Wasserstein distance, $\Gamma(\mu,\nu)$ denote all joint distributions $\mathrm{\pi}$ that have marginals $\mu$ and $\nu$, $d$ is a metric (distance function) between points $x$ and $y$ that are being matched and $M$ is a given metric space. When $p=1$ the distance is also known as the Earth Mover Distance and carries a nice intuitive interpretation of visualizing the two distributions as piles of earth (soil). The distributional departure is then measured by the work required to turn one pile into the other through an optimal transport problem. 

\subsubsection{Coverage}
Coverage may be quantified through diversity measures. We bring attention to measures evaluating either combinatorial information, called \textit{combinatorial diversity},  $C(\cdot)$ or geometric coverage, called \textit{geometric diversity}, $G(\cdot)$. To define these metrics consider a set of observations $X$ and a discrete categorical feature with $k$ categories. This gives rise to a partition of the dataspace into $k$ parts $X = X_1 \cup X_2 \cup.. \cup X_k$, leading to a combinatorial measure of diversity. The combinatorial diversity of a subset $S \subseteq X$ is defined as the Shannon entropy of the distribution \cite{celis2016fair}:
\begin{equation}
    C(S) = -\sum\limits_{i=1}^k s_i \ \mathrm{log} \ s_i,
    \label{Eq:Entropy}
\end{equation}
where the combinatorial diversity measure $C(S)$ is the Shannon entropy, $s_i = \frac{|S|\cap|X_i|}{|S|}$ is the probability of event $i$ and $\sum$ is the sum over the possible outcomes. Thus combinatorial diversity (also known as diversity index \cite{simpson1949measurement}) has roots in information theory and measures the degree of diversity through the Shannon entropy of the distribution. High entropy corresponds to high diversity. The combinatorial diversity measure is useful to quantify diversity in features with a set of discrete human-interpretable values \cite{celis2016fair} such as race.

On the other hand, geometric diversity is motivated from a volumetric perspective \cite{celis2016fair}. Each datapoint $x \in X$ is represented by a feature vector $v_x$. The geometric diversity of a subset $S \subseteq X$ is the $n$-volume of the parallelotope spanned by the $n$ feature vectors $\{v_x:x \in S\}$, where $n=|S|$ is the size of the subset. Denoting the data matrix of the subset $S$ as $\bold{D} \in \mathbb{R}^{p \times n}$, the (squared) $n$-volume of the $n$-parallelotope embedded in a $p$-dimensional space (where $p > n$) can be computed by means of the determinant of the Gramian matrix $\bold{G} = \bold{D}^T\bold{D}$ (with feature vectors as columns in $\bold{D}$). Thus the geometric diversity can be measured by:

\begin{equation}
    G(S) = \sqrt{Det(\bold{D}^T\bold{D})},
    \label{Eq:Volume}
\end{equation}
where $G(S)$ is the geometric diversity of subset $S$, $Det({\cdot})$ denotes the determinant and $\bold{D}$ is the data matrix of the subset $S$. Geometric diversity is motivated from a perspective of diverse feature vectors. Intuitively, diverse vectors can be interpreted as divergent and thus pointing in different directions. The diversity of these can be measured by the volume of the parallelotope spanned by the vectors. Thus, the larger the volume, the higher the geometric coverage. Geometric diversity is closely related to a type of probability distribution known as \textit{determinantal point process} (DPP) \cite{kulesza2012determinantal}, which can be used to draw samples proportional to their geometric diversity.

While geometric diversity can be a good measure of the coverage for a sample, or between different samples of the same size from the same population, it does not directly relate to the degree of coverage in the original population space. As the metric evaluates $n$-dimensional volumes, comparing geometric diversity between sample and population equates to comparing different dimensional volumes. On the other hand, comparing between different sized samples through combinatorial diversity is straightforward, as this measure operates intrinsically on normalized probabilities.

\subsubsection{Representatives}
A typical or ideal observation may be estimated as the mean, centroid or mode of the group it represents, and the representativeness may be measured by the variance in the group. Furthermore, in settings where representativeness of an underlying population is sought through data reconstruction from a combination of archetypes (ideal exemplars), the representativeness of these archetypes can be measured through a reconstruction loss between original and reconstructed data.

\subsubsection{Fairness measures} 
Analogous to how the notions of representativity may be measured mathematically, various measures also exist to quantify the adverse effects of insufficient representation, known as representation bias. Representation bias occurs when parts of the input space are underrepresented \cite{suresh2019framework,shahbazi2022survey}, for instance a sampled population which underrepresents and fails to generalize well for parts of the population, which can manifest as disparate predictive accuracy for these groups \cite{chen2018my,asudeh2019assessing,jin2020mithracoverage}. Common cases include models trained on ImageNet \cite{deng2009imagenet,shankar2017geodiversity} and commercial facial analysis algorithms \cite{buolamwini2018gender}. While these models are not intrinsically unfair, they may capture and increase biases present in the training data. This inherited bias can be measured through algorithmic fairness metrics. 

Algorithmic fairness is often formulated in terms of independence relations between model predictions $\hat{Y}$ and a protected attribute $A$ (typically a binary feature $A \in \{0,1\}$) denoting group membership under a protected category such as race or sex. A common notion of algorithmic fairness, known as demographic parity (or statistical parity), is defined to require independence between model predictions and a protected attribute:
\begin{equation}
    \hat{Y} \indep A
    \label{Eq:Formal_Statistical_Parity}
\end{equation}
In regression settings $\hat{Y}$ is a real-valued random variable characterized by its cumulative distribution function (CDF). The departure of the CDF of $\hat{Y}$ from the CDF of $\hat{Y}$ conditional on the protected attribute $A$ can be used as a measure of demographic parity \cite{agarwal2019fair,ruf2021implementing}. For a binary decision problem $\hat{Y} \in \{0,1\}$ with a binary protected attribute $A \in \{0,1\}$ the demographic parity constraint can be expressed by $P\{\hat{Y}=1 | A=0\} = P\{\hat{Y}=1 | A=1\}$ \cite{hardt2016equality}, thus requiring equality of positive rates for subsets of the protected attribute.

The demographic parity criterion has been critiqued on various accounts \cite{dwork2012fairness,hardt2016equality}, which has lead to an alternative formulation of algorithmic fairness known as equalized odds. Equalized odds formulates the following conditional independence:
\begin{equation}
    \hat{Y} \indep A \ | \  Y,
    \label{Eq:Formal_Equalized_Odds}
\end{equation}
The equalized odds constraint applies to targets and protected attributes in any space \cite{hardt2016equality}. For binary classification with a binary protected attribute, the constraint can be formulated as $P\{\hat{Y}=1 | A=0,Y=y\} = P\{\hat{Y}=1 | A=1,Y=y\}, y \in {0,1}$, where $y$ is the outcome. In this setting $Y=1$ is often considered the advantaged outcome, which leads to a popular relaxation of the equalized odds measure known as \textit{equal opportunity} \cite{hardt2016equality}. This measure prohibits discrimination only within the advantaged outcome group and can be formulated as $P\{\hat{Y}=1 | A=0,Y=1\} = P\{\hat{Y}=1 | A=1,Y=1\}$. Equal opportunity thus requires equality of true positive rates.  

\section{Survey of use in AI literature}
\label{sec:Survey}
To provide insight into the use of data representativity in current literature, we conduct a survey of papers from two typical and highly recognized AI conferences; The Conference on Neural Information Processing Systems (NeurIPS) and The International Conference on Computer Vision (ICCV). We restrict our survey to papers contributing novel datasets at either the NeurIPS 2021 Track on Datasets and Benchmarks or at the main conference at ICCV. The NeurIPS track has 174 accepted papers contributing either high-quality datasets, new benchmarks or discussions on data related work; 108 of them contribute novel datasets. These papers are required by NeurIPS guidelines to provide dataset documentation and intended uses. The organizers recommended using documentation such as datasheets for datasets, which encourages the authors to consider and document how their work relates to data representativity. The main conference at ICCV 2021 has 1612 accepted papers, of which we identify 32 contributing novel datasets. We conduct the survey by reviewing which notions each paper uses to describe the representativity of their dataset. A summary of the survey results can be found in Table \ref{tab:survey}.

 \begin{table}[h!]
   \caption{Summary of notions used in the 108 papers introducing novel datasets on the NeurIPS 2021 Track on Datasets and Benchmarks and 32 papers introducing novel datasets from ICCV 2021. For each paper we document which notion they use to describe the representativity of their dataset. A paper can use more than one notion of representativity.}
  \label{tab:survey}
  \begin{tabular}{lclcl}
    \toprule
     & \multicolumn{2}{c}{\textbf{NeurIPS 2021}} & \multicolumn{2}{c}{\textbf{ICCV 2021}} \\
    \textbf{Notion} & \textbf{Number of papers} & \textbf{Percent} & \textbf{Number of papers} & \textbf{Percent} \\
    \midrule
    No notion              &  2 & 1.9 \% & 1  & 3.2 \% \\
    Assertive              & 10 & 9.3 \%  & 2  & 6.3 \%\\
    Miniature              & 15 & 13.9 \%  & 10  & 31.3 \%\\
    Selective Forces       & 41 & 38.0 \%  & 7  & 21.9 \%\\
    Typical / Ideal        & 14 & 13.0 \%  & 4  & 12.5 \%\\
    Coverage               & 66 & 61.1 \%  & 27  & 84.4 \%\\
    Reference to sampling  & 108 & 100.0 \%  &  30 & 94 \%\\
    Copycat               & 18 & 16.7 \%  & 5  & 15.6 \%\\
  \bottomrule
\end{tabular}
\end{table}

\subsection{Examples}
We find that the various notions appear in a wide array of settings and range from implicit to explicit use. Here we provide noteworthy examples demonstrating how the authors use the notions to express the representativity of their datasets.

\subsubsection{Assertive claim}
The assertive claim appears in about $9\%$ of surveyed NeurIPS publications and $6\%$ of the surveyed ICCV papers. An example of the notion can be found in the datasheet of a publication regarding time-sensitive questions \cite{survey_10}: "It’s sampled from large Wikipedia passages, it’s representative of all the possible temporal-sensitive information." Further examples include \cite{Survey_47}: "We select 16k most representative scenes and exhaustively annotate all the 3D bounding boxes of 5 categories..." and \cite{Survey_69} "This data contains examples of slang, acronyms, lack of punctuation, poor orthography, concatenations, profanity, and poor grammar, among other forms of atypical language usage. This data is representative of the types of inputs that machine translation services find challenging."

\subsubsection{Miniature} 
The miniature notion appears in roughly 14 $\%$ of the surveyed NeurIPS papers and  $31\%$ of the surveyed ICCV papers. It emerges in various settings including demographic population representativity of people \cite{Survey_71}: "Tab. 1b shows a statistical summary of the eligible cohort. This cohort broadly reflects the Tufts student population in terms of age, racial and gender makeup." Likewise the notion is used in relation to population distribution of animals in a paper regarding animal pose estimation \cite{Survey_61}: "... the number of images in each family of AP-10K has a long-tail distribution, which reflects the true distribution of animals in the wild due to the commonness or rarity of the animals in some extent."

The notion also appears in more restricted forms, for instance claiming a miniature in terms of a specific geographic region \cite{Survey_36}: "The class imbalance provides a challenge for machine learning algorithms but it is representative of the geographic region and an imbalance is generally common in real-world crop type mapping tasks." Furthermore, the notion also appears under the disguise of 'representative coverage' \cite{Survey_90}: "By restricting a dataset to only those tweets matching a pre-defined vocabulary, a higher percentage of hateful content can be found. However, this sacrifices representative coverage for cost-savings, yielding a biased dataset whose distribution diverges from the real world we seek to model and to apply these models to in practice." 

Finally, some publications state that their data are not representative in terms of the miniature notion \cite{Survey_96}: "Two thirds of the dataset concentrate on as few as four countries: Germany, France, the UK, and Spain. This distribution is not representative of the actual distribution of church buildings across Europe but most likely correlated with the size and level of activity of the local Wikipedia communities and their propensity to enter information in Wikidata." 

\subsubsection{Selective forces}
The selective forces notion appears in 38 $\%$ of the surveyed NeurIPS papers and $22\%$ of the surveyed ICCV papers. The notion is mostly used to claim non-representativity due to the presence of selective forces in the sampling process. Examples include: \cite{Survey_71}: "First, our dataset is limited in whom it represents. Because we draw from a convenience sample at our university, ages are skewed toward typical college students and the racial makeup reflects that our campus community is largely white and Asian." Another example is \cite{Survey_106} "Because our dataset comprises only named and published chaotic systems, it does not comprise a representative sample of the larger space of all low-dimensional chaotic systems." Yet another example includes a discussion on the difficulty of dealing with multiple selective forces \cite{Survey_4}: "... sampling images randomly from an uncurated large collection removes specific biases such as search engine selection but not others, for example the geographic bias. Furthermore, we added one significant bias: there are no people in these pictures, despite the fact that a large fraction of all images in existence contain people"

On the other hand the absence of selective forces is used as an indication that no sampling biases exist, and hence that the data is representative of the population \cite{Survey_80}: "To avoid introducing biases, all comments of the RP have been considered without further topical filtering. Furthermore, using a broad crowd to annotate the data should minimize the inclusion of person- specific biases."

\subsubsection{Typical / Ideal}
This notion is used in 13 $\%$ of the surveyed papers and also appears in relation to synthetic data generation, where representativity of the underlying population is sought modelled through variation on ideal / archetypical patterns or shapes. For example \cite{Survey_1}: "The first part of the template specification describes a base sewing pattern that would then be parametrized and varied to produce new designs." "The training group of 12 templates aims to cover design spaces of typical simple garments, including skirts, dresses, tops, pants, jackets, hoodies, and jumpsuits, and reflect some topological variations among them."

The notion is also used to express representativity of a population through typical systems or methods \cite{Survey_29}: "We propose four representative physical systems, as well as a collection of both widely used classical time integrators and representative data-driven methods (kernel-based, MLP, CNN, nearest neighbors)".

\subsubsection{Coverage}
The coverage notion is popular appearing in 61 $\%$ of the surveyed NeurIPS papers and $84\%$ of the surveyed ICCV papers. The notion can be found in a wealth of settings and often appears as a claim of diversity in the data, for example \cite{Survey_1}: "... the motivation was to resemble the variety of designs that exist within a garment type while covering this diversity uniformly." Another example is \cite{Survey_44} "A diverse quantity of wild and lab culture mosquitoes is included in the database to capture the biodiversity of naturally occurring species." The notion is also use in terms of language coverage in \cite{Survey_108}: "The dataset is our best effort to extract and represent as much diversity (in terms of various different languages) from Common Voice as possible." Additionally, the notion is commonly used in ICCV papers to support that the published image data is representative of the real world in terms of visual diversity \cite{Survey_ICCV_1}: "...we wanted scenes that are as photorealistic and visually diverse as possible." 

\subsubsection{Copycat}
The copycat notion can be found in about 16 $\%$ of the surveyed papers and appears mostly in relation to synthetic data generators constructed to copy or mimic the distribution of real-world data. For instance \cite{Survey_74}: "The synthetic datasets we release offer a wide variety of parameters that can be configured to simulate real-world data." Another example is \cite{Survey_45}: "Note that this data captures the behavior of real workers in the target domain modulo potential differences induced by the use of a synthetic speech generator."

The notion is however also used in tandem with the notion of selective forces to deliberately synthesize data that diverges from the real-world distribution. \cite{Survey_64}: "We took advantage of the synthesis pipeline to showcase how datasets can be constructed with properties that deliberately differ from real world distributions. Notably, we include samples of individuals with common (e.g., scientist) as well as uncommon occupations (e.g., spy) (Table 3) and designed SynthBio to be more balanced with respect to gender and nationality compared to the original WikiBio dataset." 

\section{Survey Discussion}
We find that the various notions of representative samples are still highly pertinent. Over $95\%$ of the surveyed papers use at least one notion and all notions appear in a wide range of settings. Overall, we observe similar occurrence rates for the notions across the two conferences, with the largest differences apparent in the use of miniature, selective forces, and coverage notions. For both conferences the coverage notion is especially prominent appearing in $61\%$ of the surveyed NeurIPS paper and $84\%$ of the surveyed ICCV papers. This might partially be attributed to the backdrop of the cautionary tale on lack of coverage in ImageNet, bu also partly due to questions in datasheets for datasets using a clear notion of coverage (e.g. geographic coverage) when inquiring about the representativity of the dataset. We also bring attention to the the assertive notion, which is rarely used but still has a somewhat high occurrence rate considering the recognition of the two conferences. 

\section{Demonstrations using data}\label{sec:data}
To demonstrate contrasting perspectives on representativity, we empirically evaluate performance, fairness and diversity for samples created with either with the concept of coverage or reflection in mind. The samples are created from a US census data collection \citep{ding2021retiring}
through stratified random sampling to obtain a miniature and through either density based or determinantal point process (DPP) based sampling to achieve coverage. 

The US Census data exhibit significant population skew between minority and majority groups of protected attributes as well as significant interstate geographical variation. For this reason the data provides a suitable testing ground to study the effects of data representativity in relation to representation bias. Models trained on biased data can result in learned mappings from input to output that are uncertain for underrepresented regions \cite{suresh2019framework}, which may lead to disparate predictive accuracy for different groups \cite{chen2018my,asudeh2019assessing,jin2020mithracoverage}, but can also cause adverse effects on overall performance under distributional shifts between training and target data. 

For instance if models trained on specific states are applied to other states \cite{ding2021retiring}. Representation bias can be mitigated by identifying and populating underrepresented parts of the data distribution \cite{suresh2019framework,jin2020mithracoverage}. Such mitigation efforts could be performed by obtaining additional data, by targeted data augmentation (eg. SMOTE \cite{chawla2002smote}) or by probabilistic over-sampling of underrepresented data regions \cite{kjaersgaard2021sampling}. Representation bias can occur in real-world ML applications, where a systemic bias in the geographical distribution of US cohorts used to train models for clinical applications has been uncovered \citep{kaushal2020geographic}. This investigation found that 71$\%$ of the analyzed studies used cohorts from at least 1 of 3 states, namely California, Massachusetts or New York, while 34 states did not contribute to any cohorts. California cohorts appeared in 39$\%$ of all analyzed studies. With this in mind, we also investigate the role of data representativity for drawing inference under distributional shifts, by comparing performance on in-distribution and out-of-distribution data for the different sampling strategies.  

\subsection{Data}
\label{sec:Data}
The UCI Adult dataset from the 1994 Current Population Survey is organized by the US Census Bureau \citep{kohavi1996adult} and is a popular dataset in the machine learning community. This data has been used in hundreds of research papers, but its external validity has been questioned, and a collection of new datasets from US Census Bureau data have been proposed \citep{ding2021retiring}. More specifically, these datasets are extracted from the American Community Survey Public Use Microdata Sample (ACS PUMS). They contain data on attributes like age, income, education, sex, ancestry and employment. The responses to the survey are controlled by privacy rules seeking to prevent re-identification of responders. Detailed documentation on the records can be found on the US Census Bureau websites. One of the proposed datasets is a replacement for the original UCI Adult dataset containing an income prediction task for a feature subset of the 2018 ACS PUMS data spanning all US states in addition to Puerto Rico.
\begin{table}[h!]
  \caption{Overview of the features in the modified US Census income data (n=1,664,500). Features COW, SCHL, MAR, POBP and RELP are modified from the original ACS PUMS data by binarizing into respectively government / non-government worker (COW), Bachelor's degree / no Bachelor's degree (SCHL), married / not married (MAR), US-born / non-US-born (POBP) and reference person / non-reference person (RELP). See the ACS PUMS dictionary documentation for full feature descriptions including original category codes.}
  \label{tab:income_features}
  \begin{tabular}{cccccl}
    \toprule
    Feature Type & Feature Name & Description & Data Type & Categories & Min/Max\\
    \midrule
    Input & AGEP & Age & Continuous & - & 17 - 96 \\
    Input &COW & Class of worker & Binary & 2 & - \\
    Input &SCHL & Educational attainment & Binary & 2 & - \\
    Input &MAR & Marital status & Binary &2 & -\\
    Input &POBP & Place of birth & Binary & 2 & -\\
    Input &RELP & Relationship & Binary & 2 & -\\
    Input &WKHP & Hours worked per week & Continuous & - & 1 - 99 \\
    Input &SEX & Sex & Binary & 2 & - \\
    Input &RAC1P & Race & Categorical & 9 & -\\
    Target &PINCP & Total income & Continuous & - & 104 - 1,423,000  \\
  \bottomrule
\end{tabular}
\end{table}

To generate the dataset the ACS PUMS data are filtered to only include individuals over the age of 16 with at least one working hour per week and an income of at least 100 USD in the past year. This leaves a total of 1,664,500 individuals. Like the original UCI Adult dataset, this new dataset has a predefined income threshold of 50,000 USD used to binarize the targets into a classification setting. Fairness intervention tasks have been shown to be sensitive to the specific threshold value \citep{ding2021retiring}. For this reason we create a modified version of the income dataset and omit the income threshold to form a regression task with the continuous income as target. We transform the income target using the natural logarithm to obtain homoscedasticity for the residuals in our regression model. An overview of the dataset can be seen in Table \ref{tab:income_features}. 

\subsection{Methodology}
We compare linear regression models fitted to the log transformed income using all features in Table \ref{tab:income_features} for the state of California (n=195,665). We evaluate model performances using 5-fold cross validation where for each iteration $20\%$  (n=39,133) of the California data are used for testing and the remaining $80\%$ (n=156,532) are used for training. We compare a model trained using the full training data (which we denote full census model) to models trained on samples of the training data following either the reflection or coverage concepts of representativity. For each iteration a miniature and coverage sample is drawn from the training data. The sample sizes are $20\%$ (n=31,306) of the full census training data. We evaluate the concepts of representativity by comparing performances on a range of metrics including overall performance using the mean squared errors (MSE) as well as performance in terms of fairness and diversity criteria. We also evaluate performance on in-distribution and out-of-distribution data by comparing interstate and intrastate performance. For completeness we show additional results from logistic regression classification models on the original binarized income (50,000 USD threshold) in Appendix \ref{Append:CA}.

\subsubsection{Generating Samples}
We generate miniature samples using a population based probability sampling scheme known as proportional stratified random sampling. Based on various demographic features the data are subdivided into smaller groups (strata) that exhibit homogeneity. Subsequently random samples are drawn from these strata. To ensure such sampling constitutes a true miniature of the underlying population requires either a relatively homogeneous population or an increasingly large sample the more sociodemographic features are considered. This is particularly the case with sociodemographic data containing minority groups, where strata can become too finely grained and be represented by statistically insufficient sample sizes \cite{bornstein2013sampling}. To sample rare ethnic groups disproportionate sampling (for instance oversampling of minority groups) can be used \cite{kalsbeek2003sampling,kalsbeek2007disproportionate,chen2021oversampling}, but this can lead to adverse affects on overall population estimates. To avoid too finely grained strata we generate miniature samples by cross stratifying on three important protected sociodemographic features, namely age, sex and race. The sex feature contains 2 categories, while the race feature contains 9 categories. We bin the age feature into three bins containing age groups of [0-33],[33-66],[66-99]. This combines to a total of 54 strata. In section \ref{sec:results} we empirically demonstrate that our stratified random sampling mimics the population and that results on the miniature samples generalize to the population.

We generate coverage samples using two approaches. Firstly, a density based coverage approach using density weighted sampling proposed in \citep{kjaersgaard2021sampling}. The density around observations is measured by the mean distance to the nearest neighbors and the density measures are then used as sampling probabilities in a weighted random sampling scheme. This approach causes observations in low-density regions to be sampled with high probability and conversely observations from high-density regions to be sampled with low probability. In doing so, the density sampling equally covers the input space regardless of the demographic proportions in the population. 

Secondly, we generate a diverse coverage sample using a determinantal point process (DPP) probability distribution \cite{kulesza2012determinantal}. DPPs have been used to create diverse sets in a number of ML applications ranging from documents, sensors, videos, images and recommendations systems. \cite{lin2012learning,krause2008near,gong2014diverse,kulesza2012determinantal,zhou2010solving}. The DPP is a distribution over subsets $S$ such that the probability of a subset is proportional to the determinant of a positive semidefinite kernel matrix known as the L-ensemble $P(S) \propto Det(L)$. The L-ensemble may be constructed as the Gramian of the data. Since inference through DPPs rely on inversion and eigendecomposition of the L-ensemble, this procedure is inefficient with large $N$, where typically the dual representation is used for efficient inference over large sets \cite{kulesza2012determinantal}. DPPs model not only the content of the subsets, but also the size. To draw samples of a specific size k-DPPs, a conditional DPP modeling only subsets of cardinatliy k, was proposed \cite{kulesza2011k-dpp}. We generate our DPP samples with the DPPy library \cite{gautier2019dppy} using k-DPPs through the dual representation.

\subsubsection{Out-of-distribution performance}
To investigate the role of data representativity for drawing inference under distributional shifts, we compare performance on in-distribution (the California test data) and out-of-distribution data (the remaining 49 US states and Puerto Rico) for the different sampling strategies.  

\subsubsection{Fairness metrics}
We measure group level fairness between the overrepresented group of White individuals (accounting for 62.2$\%$ of the California data) and the underrepresented group of Native American individuals (accounting for 0.9$\%$ of the California data). In the ACS PUMS data Native Americans include both American Indian and Alaska Native individuals. We measure fairness based on \textit{demographic parity} and \textit{equalized odds} defined in Equations \ref{Eq:Formal_Statistical_Parity} and \ref{Eq:Formal_Equalized_Odds}. We quantify demographic parity for our regression models by measuring the departure of the CDF of model predictions to the CDF of model predictions conditional on the protected attribute. We denote this departure the regression demographic disparity (RDD) and measure it using the $\ell_{\infty}$ norm. We measure the equalized odds disparity using an approach based on resampling of protected attributes \cite{romano2020achieving}. Here a synthetic resampled version of $A$ is constructed, called fair dummies $\tilde{A}$, such that the triple $(\hat{Y},\tilde{A},Y )$ obeys equalized odds. The distribution of the fair triple $(\hat{Y},\tilde{A},Y)$ is then compared to that of the observed test data $(\hat{Y},A,Y)$. We again measure the distributional departure using the $\ell_{\infty}$ norm and denote this the regression equalized odds disparity (REOD). For our classification models we measure fairness in terms of demographic parity and equal opportunity by the difference in positive rates and the difference in true positive rates between White and Native American individuals. We denote these measures the classification demographic disparity (CDD) and classification equal opportunity disparity (CEOD).

\subsubsection{Coverage Metrics}
\label{sec:method_diversity}
We compare samples on combinatorial diversity $C(\cdot)$ and geometric coverage $G(\cdot)$ defined in Eqs. \ref{Eq:Entropy} and \ref{Eq:Volume}. Typically geometric coverage is computed from the determinant of the L-ensemble (Gramian), but for the US Census data $p \ll n$, which leads to a determinant and volume of zero. This necessitates an alternative formulation. We instead compute the diversity from the dual representation of the L kernel, which carries information about several important properties of the L-ensemble \cite{DPPforML}. 

\subsubsection{Reflection Metrics}
We evaluate the reflection concept of representativity for the samples both through statistical tests on average predictions between sample and population, as well as a measure of distance between overall sample and population distributions. For the distributional measure we report the first Wasserstein distance between samples and population for two features.

\subsection{Results}
\label{sec:results}

The MSE on the in-distibution California test data can be seen in Table \ref{tab:results_regression}. The model trained on the full census training data has the lowest MSE followed by the miniature and DPP model, while the density model has the highest MSE. Table \ref{tab:results_regression} also illustrates how the models score on fairness criteria for demographic parity and equalized odds between White and Native American individuals. The density model has the best performance in terms of demographic parity and equalized odds while the miniature and full census model have the worst performances. P-values from paired t-tests on sample results can be found in Appendix \ref{Append:CA} in Table \ref{tab:t-test_regression}. Equivalent results for the classification setting can be found in Appendix \ref{Append:CA} in Tables \ref{tab:results_classification} and \ref{tab:t-test_classification}.

 \begin{table}[h!]
   \caption{Regression performance metrics evaluated with 5-fold cross validation on the California data. For each iteration 80$\%$ (n=156,532) of the California data is used for training a full census model and the remaining 20$\%$ (n=39,133) is used for testing. For each iteration a miniature and coverage sample is drawn from the full census training data and tested on the test data. The sample sizes are 20$\%$ (n=31,306) of the full census training data. The overall MSE across the 5 folds is shown in addition to regression demographic disparity (RDD) and equalized odds disparity (REOD) for White / Native American individuals.}
  \label{tab:results_regression}
  \begin{tabular}{ccccccccl}
    \toprule
    Training Data & MSE & Parity (RDD) & Equality (REOD) & MSE SD & Parity SD & Equality SD  \\
    \midrule
    Full Census      & \textbf{0.7912} & 0.2286   &           0.0016 & 0.0103 & 0.0167 &  0.0002 \\
    Miniature Sample &  0.7915         & 0.2401   &           0.0017 & 0.0102 & 0.0103 &  0.0002\\
    Density Sample   &  0.8321 & \textbf{0.1637}  &  \textbf{0.0010} & 0.0092 & 0.0097 &  0.0003\\
    DPP Sample       &  0.7989 &         0.2144   &           0.0015 & 0.0085 & 0.0135 &  0.0001\\
  \bottomrule
\end{tabular}
\end{table}

We report sample scores in terms of their combinatorial (Eq. \ref{Eq:Entropy}) and geometric (Eq. \ref{Eq:Volume}) diversity in Table \ref{tab:CA diversity}.

 \begin{table}[h!]
 \caption{Mean combinatorial diversity $C(\cdot)$ on the race feature and geometric diversity $G(\cdot)$ on all features for the five samples of each sample type. Standard deviations (SD) are also shown.} 
  \label{tab:CA diversity}
  \begin{tabular}{cccccl}
    \toprule
    Sample Type & \textit{$C(\cdot)$}  & \textit{$G(\cdot)$} & \textit{$C(\cdot)$} SD & \textit{$G(\cdot)$} SD \\
    \midrule
    Miniature    & 1.184            & $0.014 \cdot 10^{27}$ & 0.001  & $0.001 \cdot 10^{27}$    \\
    Density      & 1.820            & $3.584 \cdot 10^{27}$ & 0.004  & $0.203 \cdot 10^{27}$  \\
    DPP          & \textbf{1.939}   & $\mathbf{26.100 \cdot 10^{27}}$ & 0.001 & $0.383 \cdot 10^{27}$    \\
  \bottomrule
\end{tabular}
\end{table}

Table \ref{tab:CA miniature} reports distributional distances to asses the reflection concept of representativity for the different samples. 
 \begin{table}[h!]
 \caption{Distributional comparisons to evaluate the reflection concept of representativity. Comparisons are reported as mean 1D Earth Mover Distances (EMD) between full census and the respective samples of each type. Standard deviations (SD) are also shown. EMD between population and sample is reported for the race and hours worked feature. The miniature samples where created by stratifying on the race feature, but not on the hours worked feature. Lower distance equates to higher evidence toward the reflection concept of representativity.} 
  \label{tab:CA miniature}
  \begin{tabular}{cccccl}
    \toprule
    Sample Type & EMD Race & EMD Hours Worked & Race SD & Hours Worked SD \\
    \midrule
    Miniature     & \textbf{0.000} & \textbf{0.066} & 0.000  & 0.007   \\
    Density       & 1.318          & 3.709          & 0.016  & 0.035   \\
    DPP           & 0.817          & 1.151          & 0.007  & 0.049   \\

  \bottomrule
\end{tabular}
\end{table}

\subsubsection{Out-of-distribution results}
\begin{figure}[h!]
    \centering
    \begin{subfigure}[b]{0.45\textwidth}
        \includegraphics[width=1\textwidth]{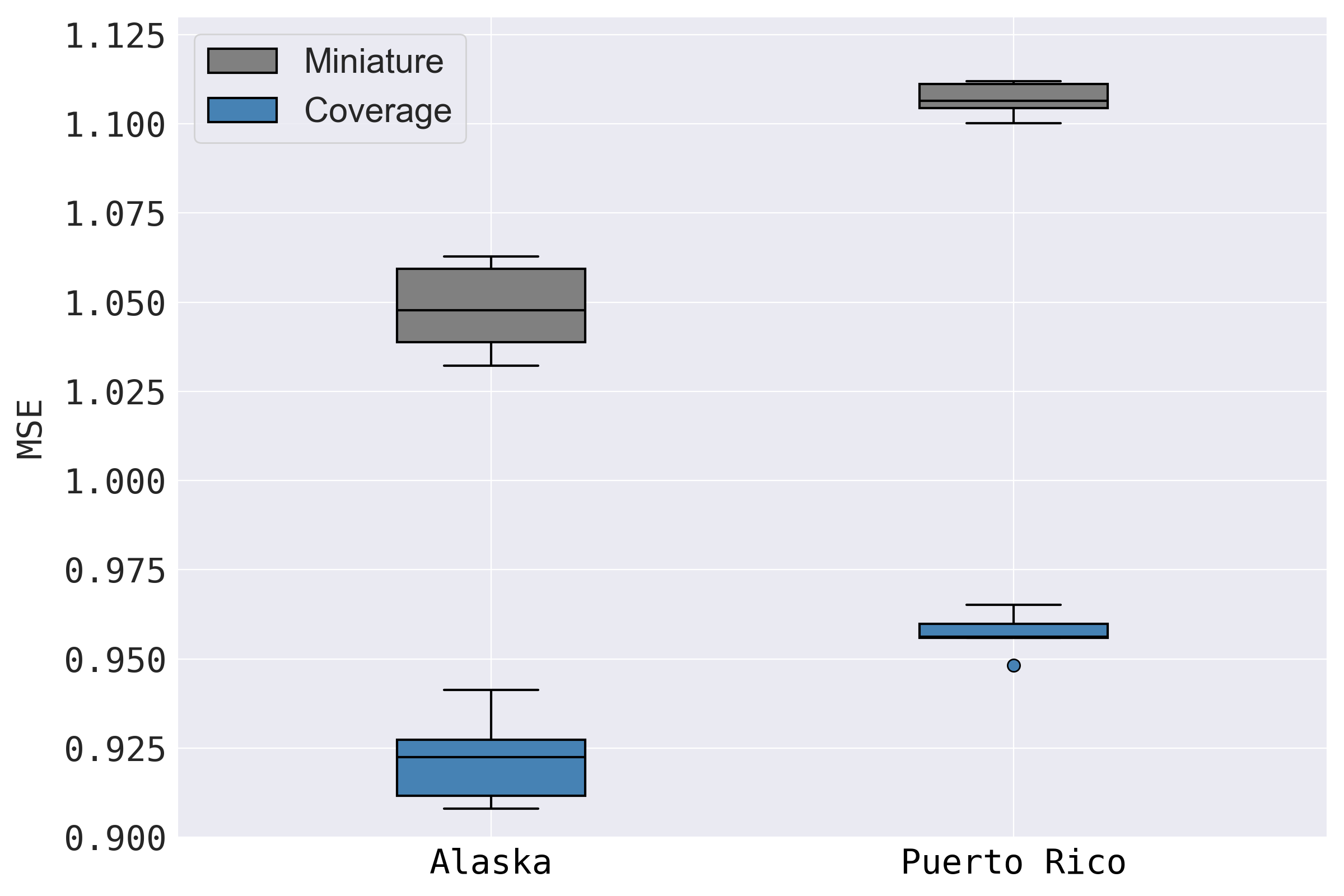}
        \caption{MSE on states dissimilar to the training data (California).}
        \label{fig:Robust_ca1}
    \end{subfigure}
    \vspace{\baselineskip}
    \begin{subfigure}[b]{0.45\textwidth}
        \includegraphics[width=1\textwidth]{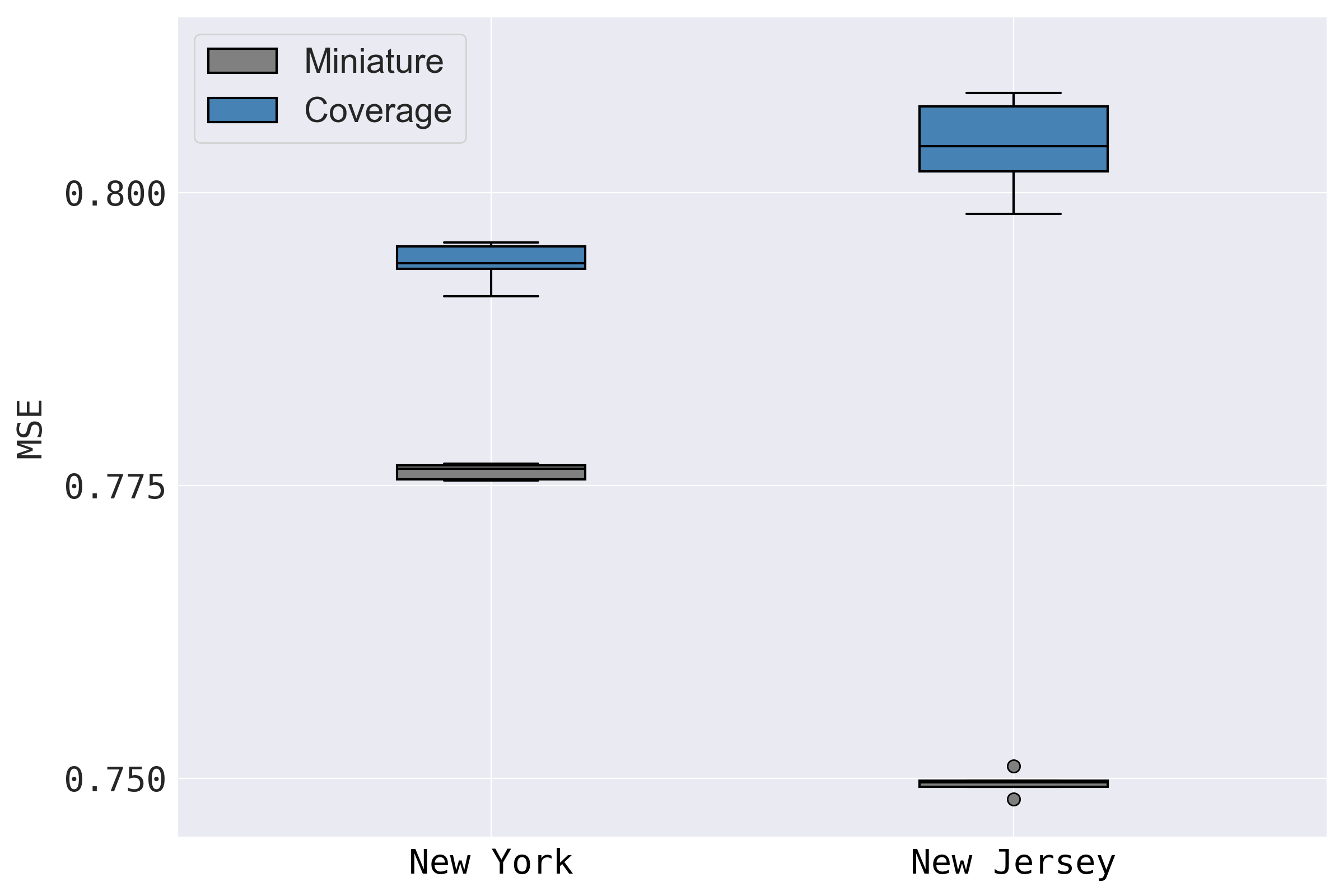}
        \caption{MSE on states similar to the training data (California).}
        \label{fig:Robust_ca2}
    \end{subfigure}
    \caption{MSE performance when applying models trained on the California data to states either demographically dissimilar (a) or similar (b) to the training data in terms of Pearson product-moment correlation between all features. The miniature model is trained on stratified random samples, while the coverage model is trained on density samples.}\label{fig:CA_MSE_OOD_box}
    \Description[Visualization of MSE performance on in-distribution and out-of-distribution test data for regression models trained on miniature and coverage samples of the California data.]{The figure compares models trained on miniature and coverage samples of the California data on two states with similar demographic distribution to California (New York and New Jersey) and two states with dissimilar demographic distribution (Alaska and Puerto Rico). It can be seen that the model trained on coverage samples is best on out-of-distribution test data (Alaska and Puerto Rico), but performs worse on states with similar distribution to the training data (New York and New Jersey).}
\end{figure}

We demonstrate out-of-distribution performance by applying models trained on California to the remaining 49 states and Puerto Rico. Figure \ref{fig:CA_MSE_OOD_box} compares MSE performance of miniature and density coverage models on two states similar and two states dissimilar to the California training data in terms of demographic distribution. See Fig. \ref{fig:CA_Regression} in Appendix \ref{Append:CA} for an out-of-distribution performance breakdown on the remaining states. MSE performance on in-distribution data is best for the model trained on miniature samples of the California training data, while MSE performance on out-of-distribution data is best for the model trained on coverage samples of the California training data. Overall the model trained on density coverage samples is on average better on 41 of the 50 states and Puerto Rico with an average performance increase of 4$\%$ across all states. Similar results can be found for the classification case in Appendix \ref{Append:CA}, where the coverage model is better than the miniature model on 43 of the 50 states and Puerto Rico with an average accuracy increase of 1.5$\%$.
Fig. \ref{fig:MA_Regression} in Appendix \ref{Append:MA} shows results of models trained on a state with different demographic distribution than California. Here we use Massachusetts as training data and again find a model trained on miniature samples to achieve better predictive performance on in-distribution data, but worse performance on out-of-distribution data.

\subsection{Summing up experiments on data}
While the coverage sampling has merits such as robustness to distributional shifts and less disparate predictive performance between under- and overrepresented parts of the input space, the coverage sampling fails to accurately represent the distribution of the underlying population, and consequently incurs a loss in predictive power on the majority of said population, measured by the MSE. On the contrary, the miniature sampling accurately represents the underlying demographic distribution of the population allowing a similar interpretation of relations between sample and population. Consequently the miniature sampling is particularly appropriate for historical or in-distribution inference on the majority. This is evident for model performances on in-distribution data, where the miniature sampling achieves better predictive performance than the coverage sample.

While we demonstrate improved race representation for our coverage sampling and consequently less disparate predictive accuracy between these groups, it should be noted that coverage procedures cannot be blindly applied to any dataset with the expectation of improved representation for marginalized groups. For instance \cite{celis2016fair} shows that sampling for diverse image summaries with the notion of geometric coverage (DPP sampling) does not necessarily result in the desired improvement in gender representation in the generated summaries. This happens when instances of overrepresented and marginalized groups are not geometrically distinct (for instance with visually similar images of individuals of different race and gender). Likewise, the density based sampling approach relies on marginalized groups being positioned in low-density regions of the input space in order to achieve sufficient coverage of these. This underlines a point that achieving a representative sample under the concept of coverage should be seen in the context of the dataset and task at hand. Improved techniques for identifying and achieving optimal coverage of marginalized groups or regions in datasets provides an important future research direction.

\section{Framework for data representativity} \label{sec:framework}
This section presents our proposed framework of questions for assessing data representativity when creating and documenting data. The framework naturally fits into both datasheets for datasets \citep{Gebru2021} as well as shorter, more general data descriptions, and our aim here is to make it as concise and manageable as possible. With this in mind, and based on our literature study, proposed concepts, and empirical investigations, we propose answering and adhering to the following questions and guidelines:

\subsection{Purpose:} What is the purpose of collecting/creating the data, and what/who is the target population? In addition, when building AI systems; what is the intended aim of the AI system along side its intended use?

\subsection{Sampling methodology:} Which data representativity concept have you used to create your sample (reflection/coverage/representatives)? What is the sampling method and procedure used to create the data? The methodology should be specified to a degree that makes it reproducible. If a code base is used to create the data, we recommend making it open source.

\subsection{Evaluation:} Are the collected data representative of the target population or 'good enough' for the aim? We recommend making this evaluation in accordance with the purpose and measurable data representativity concept, and not as a general statement of representativity. In addition, known limitations of the representativity, in terms of coverage as well as distributional match to target population, are always desirable to document for datasets to assess possible limitations, and not least because open source datasets may be used for purposes not originally anticipated. Finally, add measures of representativity in accordance with the sampling concept and to the extend possible.

\section{Discussion} \label{sec:discussion}
We found that the notions of what constitutes a 'representative sample' from the 1979 reviews by Mosteller and Kruskal are still pertinent. When building machine learning models and AI systems, particularly two contrasting views of representativity are of relevance: The concept of coverage vs. that of a reflection. We find that the two are useful for different purposes. Coverage is useful for robustness towards distribution shifts as well as mitigation of disparate predictive accuracy between overrepresented and marginalized groups. 

The reflection concept is useful to mimic the target population allowing a similar interpretation of relations between sample and population as well as to obtain minimum average errors on the target population. However, we should keep in mind that average errors indicate that predictions are best for the majority, and not necessarily equal for population subgroups.

The notion of a 'representative sample' as an assertive acclaim without specification was mainly used in AI related literature as an implicit acclaim, without explicit mentioning of representativity, but with an indication of an inference link (generalization from data) matching that of representativity. We call for attention on such implicit use, and recommend avoiding it, thus always specifying the sampling methodology as well as the purpose and target population of the collected data along with an evaluation of representativity and limits of same for the given sample. Such specification and evaluation is critical on the path towards fully transparent and trustworthy AI systems.

Through our investigations we found that we cannot talk about general representativeness of a sample, but need to consider data collection and representativeness in coherence with our purpose (and data analysis) whether this is a research hypothesis or an aim for our AI system.

As we reach limitations from our understanding of the target distributions and/or from a large number of attributes (and their interactions), it is practically impossible to make guarantees of representativeness. As a consequence, evaluations based on several datasets as well as 'in use' data (for deployed ML models or AI systems) are encouraged. Furthermore, accounting for all possible distribution shifts that may happen in the future (where our AI system will be in production), is also practically impossible. As an alternative, or rather addition, we suggest to perform continuous monitoring of AI systems and their performance while they are in production. An AI system may also at first be deployed in shadow mode if risks are too high to use predictions without further (live) testing.

Finally, we propose that further research into measurable concepts of data representativity is necessary. There is a need for measures that are computationally feasible for large high dimensional data and which can model joint distributions (parametric and non-parametric) as well as a need for further analysis into existing measures and their limitations.

\section{Acknowledgements}
The authors would like to acknowledge colleague Murat Kulahci for insightful feedback on the manuscript. Additionally, author Rune D. Kjærsgaard is funded by a university alliance scholarship between UiB (University of Bergen) and DTU (Technical University of Denmark).

\bibliographystyle{ACM-Reference-Format}
\bibliography{sample-base}


\begin{thebibliography}{110}

%%% the Web address.

\ifx \showCODEN    \undefined \def \showCODEN     #1{\unskip}     \fi
\ifx \showDOI      \undefined \def \showDOI       #1{#1}\fi
\ifx \showISBNx    \undefined \def \showISBNx     #1{\unskip}     \fi
\ifx \showISBNxiii \undefined \def \showISBNxiii  #1{\unskip}     \fi
\ifx \showISSN     \undefined \def \showISSN      #1{\unskip}     \fi
\ifx \showLCCN     \undefined \def \showLCCN      #1{\unskip}     \fi
\ifx \shownote     \undefined \def \shownote      #1{#1}          \fi
\ifx \showarticletitle \undefined \def \showarticletitle #1{#1}   \fi
\ifx \showURL      \undefined \def \showURL       {\relax}        \fi
% invisible to TeX
\providecommand\bibfield[2]{#2}
\providecommand\bibinfo[2]{#2}
\providecommand\natexlab[1]{#1}
\providecommand\showeprint[2][]{arXiv:#2}

\bibitem[\protect\citeauthoryear{Agarwal, Dud{\'\i}k, and Wu}{Agarwal
  et~al\mbox{.}}{2019}]%
        {agarwal2019fair}
\bibfield{author}{\bibinfo{person}{Alekh Agarwal}, \bibinfo{person}{Miroslav
  Dud{\'\i}k}, {and} \bibinfo{person}{Zhiwei~Steven Wu}.}
  \bibinfo{year}{2019}\natexlab{}.
\newblock \showarticletitle{Fair regression: Quantitative definitions and
  reduction-based algorithms}. In \bibinfo{booktitle}{\emph{International
  Conference on Machine Learning}}. PMLR, \bibinfo{pages}{120--129}.
\newblock


\bibitem[\protect\citeauthoryear{Asano, Rupprecht, Zisserman, and
  Vedaldi}{Asano et~al\mbox{.}}{2021}]%
        {Survey_4}
\bibfield{author}{\bibinfo{person}{Yuki~M Asano}, \bibinfo{person}{Christian
  Rupprecht}, \bibinfo{person}{Andrew Zisserman}, {and} \bibinfo{person}{Andrea
  Vedaldi}.} \bibinfo{year}{2021}\natexlab{}.
\newblock \showarticletitle{PASS: An ImageNet replacement for self-supervised
  pretraining without humans}.
\newblock \bibinfo{journal}{\emph{arXiv preprint arXiv:2109.13228}}
  (\bibinfo{year}{2021}).
\newblock


\bibitem[\protect\citeauthoryear{Assenmacher, Niemann, M{\"u}ller, Seiler,
  Riehle, and Trautmann}{Assenmacher et~al\mbox{.}}{2021}]%
        {Survey_80}
\bibfield{author}{\bibinfo{person}{Dennis Assenmacher}, \bibinfo{person}{Marco
  Niemann}, \bibinfo{person}{Kilian M{\"u}ller}, \bibinfo{person}{Moritz
  Seiler}, \bibinfo{person}{Dennis~M Riehle}, {and} \bibinfo{person}{Heike
  Trautmann}.} \bibinfo{year}{2021}\natexlab{}.
\newblock \showarticletitle{RP-Mod\&RP-Crowd: Moderator-and Crowd-Annotated
  German News Comment Datasets.}. In \bibinfo{booktitle}{\emph{NeurIPS Datasets
  and Benchmarks}}.
\newblock


\bibitem[\protect\citeauthoryear{Asudeh, Jin, and Jagadish}{Asudeh
  et~al\mbox{.}}{2019}]%
        {asudeh2019assessing}
\bibfield{author}{\bibinfo{person}{Abolfazl Asudeh}, \bibinfo{person}{Zhongjun
  Jin}, {and} \bibinfo{person}{HV Jagadish}.} \bibinfo{year}{2019}\natexlab{}.
\newblock \showarticletitle{Assessing and remedying coverage for a given
  dataset}. In \bibinfo{booktitle}{\emph{2019 IEEE 35th International
  Conference on Data Engineering (ICDE)}}. IEEE, \bibinfo{pages}{554--565}.
\newblock


\bibitem[\protect\citeauthoryear{Barz and Denzler}{Barz and Denzler}{2021}]%
        {Survey_96}
\bibfield{author}{\bibinfo{person}{Bj{\"o}rn Barz} {and}
  \bibinfo{person}{Joachim Denzler}.} \bibinfo{year}{2021}\natexlab{}.
\newblock \showarticletitle{Wikichurches: A fine-grained dataset of
  architectural styles with real-world challenges}.
\newblock \bibinfo{journal}{\emph{arXiv preprint arXiv:2108.06959}}
  (\bibinfo{year}{2021}).
\newblock


\bibitem[\protect\citeauthoryear{Benjamini and Hochberg}{Benjamini and
  Hochberg}{1995}]%
        {benjamini1995controlling}
\bibfield{author}{\bibinfo{person}{Yoav Benjamini} {and} \bibinfo{person}{Yosef
  Hochberg}.} \bibinfo{year}{1995}\natexlab{}.
\newblock \showarticletitle{Controlling the false discovery rate: a practical
  and powerful approach to multiple testing}.
\newblock \bibinfo{journal}{\emph{Journal of the Royal statistical society:
  series B (Methodological)}} \bibinfo{volume}{57}, \bibinfo{number}{1}
  (\bibinfo{year}{1995}), \bibinfo{pages}{289--300}.
\newblock


\bibitem[\protect\citeauthoryear{Bereswicz}{Bereswicz}{2017}]%
        {Beresewicz2017}
\bibfield{author}{\bibinfo{person}{Maciej Bereswicz}.}
  \bibinfo{year}{2017}\natexlab{}.
\newblock \showarticletitle{A Two-Step Procedure to Measure Representativeness
  of Internet Data Sources}.
\newblock \bibinfo{journal}{\emph{International Statistical Review}}
  \bibinfo{number}{85} (\bibinfo{year}{2017}), \bibinfo{pages}{473--493}.
\newblock
Issue 3.


\bibitem[\protect\citeauthoryear{Blatchford, Mannaerts, and Zeng}{Blatchford
  et~al\mbox{.}}{2021}]%
        {Blatchford2021}
\bibfield{author}{\bibinfo{person}{Megan~L. Blatchford},
  \bibinfo{person}{Chris~M. Mannaerts}, {and} \bibinfo{person}{Yijian Zeng}.}
  \bibinfo{year}{2021}\natexlab{}.
\newblock \showarticletitle{Determining representative sample size for
  validation of continuous, large continental remote sensing data}.
\newblock \bibinfo{journal}{\emph{International Journal of Applied Earth
  Observations and Geoinformation}} \bibinfo{number}{94}
  (\bibinfo{year}{2021}), \bibinfo{pages}{102235}.
\newblock


\bibitem[\protect\citeauthoryear{Bornstein, Jager, and Putnick}{Bornstein
  et~al\mbox{.}}{2013}]%
        {bornstein2013sampling}
\bibfield{author}{\bibinfo{person}{Marc~H Bornstein}, \bibinfo{person}{Justin
  Jager}, {and} \bibinfo{person}{Diane~L Putnick}.}
  \bibinfo{year}{2013}\natexlab{}.
\newblock \showarticletitle{Sampling in developmental science: Situations,
  shortcomings, solutions, and standards}.
\newblock \bibinfo{journal}{\emph{Developmental review}} \bibinfo{volume}{33},
  \bibinfo{number}{4} (\bibinfo{year}{2013}), \bibinfo{pages}{357--370}.
\newblock


\bibitem[\protect\citeauthoryear{Boyd and Crawford}{Boyd and Crawford}{2011}]%
        {Boyd2011}
\bibfield{author}{\bibinfo{person}{Danah Boyd} {and} \bibinfo{person}{Kate
  Crawford}.} \bibinfo{year}{2011}\natexlab{}.
\newblock \showarticletitle{Six Provocations for Big Data}.
\newblock \bibinfo{journal}{\emph{A Decade in Internet Time: Symposium on the
  Dynamics of the Internet and Society}} (\bibinfo{date}{September}
  \bibinfo{year}{2011}).
\newblock


\bibitem[\protect\citeauthoryear{Buolamwini and Gebru}{Buolamwini and
  Gebru}{2018a}]%
        {Buolamwini}
\bibfield{author}{\bibinfo{person}{J. Buolamwini} {and} \bibinfo{person}{T.
  Gebru}.} \bibinfo{year}{2018}\natexlab{a}.
\newblock \showarticletitle{Gender Shades: Intersectional Accuracy Disparities
  in Commercial Gender Classification}.
\newblock \bibinfo{journal}{\emph{Conference on Fairness, Accountability, and
  Transparency; Proceedings of Machine Learning Research}}
  \bibinfo{volume}{81} (\bibinfo{year}{2018}), \bibinfo{pages}{1--15}.
\newblock


\bibitem[\protect\citeauthoryear{Buolamwini and Gebru}{Buolamwini and
  Gebru}{2018b}]%
        {buolamwini2018gender}
\bibfield{author}{\bibinfo{person}{Joy Buolamwini} {and}
  \bibinfo{person}{Timnit Gebru}.} \bibinfo{year}{2018}\natexlab{b}.
\newblock \showarticletitle{Gender shades: Intersectional accuracy disparities
  in commercial gender classification}. In \bibinfo{booktitle}{\emph{Conference
  on fairness, accountability and transparency}}. PMLR,
  \bibinfo{pages}{77--91}.
\newblock


\bibitem[\protect\citeauthoryear{Celis, Deshpande, Kathuria, and Vishnoi}{Celis
  et~al\mbox{.}}{2016}]%
        {celis2016fair}
\bibfield{author}{\bibinfo{person}{L~Elisa Celis}, \bibinfo{person}{Amit
  Deshpande}, \bibinfo{person}{Tarun Kathuria}, {and}
  \bibinfo{person}{Nisheeth~K Vishnoi}.} \bibinfo{year}{2016}\natexlab{}.
\newblock \showarticletitle{How to be fair and diverse?}
\newblock \bibinfo{journal}{\emph{arXiv preprint arXiv:1610.07183}}
  (\bibinfo{year}{2016}).
\newblock


\bibitem[\protect\citeauthoryear{Chawla, Bowyer, Hall, and Kegelmeyer}{Chawla
  et~al\mbox{.}}{2002}]%
        {chawla2002smote}
\bibfield{author}{\bibinfo{person}{Nitesh~V Chawla}, \bibinfo{person}{Kevin~W
  Bowyer}, \bibinfo{person}{Lawrence~O Hall}, {and} \bibinfo{person}{W~Philip
  Kegelmeyer}.} \bibinfo{year}{2002}\natexlab{}.
\newblock \showarticletitle{SMOTE: synthetic minority over-sampling technique}.
\newblock \bibinfo{journal}{\emph{Journal of artificial intelligence research}}
   \bibinfo{volume}{16} (\bibinfo{year}{2002}), \bibinfo{pages}{321--357}.
\newblock


\bibitem[\protect\citeauthoryear{Chen, Johansson, and Sontag}{Chen
  et~al\mbox{.}}{2018}]%
        {chen2018my}
\bibfield{author}{\bibinfo{person}{Irene Chen}, \bibinfo{person}{Fredrik~D
  Johansson}, {and} \bibinfo{person}{David Sontag}.}
  \bibinfo{year}{2018}\natexlab{}.
\newblock \showarticletitle{Why is my classifier discriminatory?}
\newblock \bibinfo{journal}{\emph{Advances in neural information processing
  systems}}  \bibinfo{volume}{31} (\bibinfo{year}{2018}).
\newblock


\bibitem[\protect\citeauthoryear{Chen, Stubblefield, and Stoner}{Chen
  et~al\mbox{.}}{2021a}]%
        {chen2021oversampling}
\bibfield{author}{\bibinfo{person}{Sixia Chen}, \bibinfo{person}{Alexander
  Stubblefield}, {and} \bibinfo{person}{Julie~A Stoner}.}
  \bibinfo{year}{2021}\natexlab{a}.
\newblock \showarticletitle{Oversampling of minority populations through
  dual-frame surveys}.
\newblock \bibinfo{journal}{\emph{Journal of survey statistics and
  methodology}} \bibinfo{volume}{9}, \bibinfo{number}{3}
  (\bibinfo{year}{2021}), \bibinfo{pages}{626--649}.
\newblock


\bibitem[\protect\citeauthoryear{Chen, Wang, and Wang}{Chen
  et~al\mbox{.}}{2021b}]%
        {survey_10}
\bibfield{author}{\bibinfo{person}{Wenhu Chen}, \bibinfo{person}{Xinyi Wang},
  {and} \bibinfo{person}{William~Yang Wang}.} \bibinfo{year}{2021}\natexlab{b}.
\newblock \showarticletitle{A dataset for answering time-sensitive questions}.
\newblock \bibinfo{journal}{\emph{arXiv preprint arXiv:2108.06314}}
  (\bibinfo{year}{2021}).
\newblock


\bibitem[\protect\citeauthoryear{Cornfield, Haenszel, Hammond, Lilienfeld,
  Shimkin, , and Wynder}{Cornfield et~al\mbox{.}}{2009}]%
        {Cornfield2009}
\bibfield{author}{\bibinfo{person}{Jerome Cornfield}, \bibinfo{person}{William
  Haenszel}, \bibinfo{person}{E.~Cuyler Hammond}, \bibinfo{person}{Abraham~M.
  Lilienfeld}, \bibinfo{person}{Michael~B. Shimkin}, \bibinfo{person}{}, {and}
  \bibinfo{person}{Ernst~L. Wynder}.} \bibinfo{year}{2009}\natexlab{}.
\newblock \showarticletitle{Smoking and lung cancer: recent evidence and a
  discussion of some questions}.
\newblock \bibinfo{journal}{\emph{International Journal of Epidemiology}}
  \bibinfo{number}{38} (\bibinfo{year}{2009}), \bibinfo{pages}{1175–1191}.
\newblock


\bibitem[\protect\citeauthoryear{Cutler and Breiman}{Cutler and
  Breiman}{1994}]%
        {AA1994}
\bibfield{author}{\bibinfo{person}{Adele Cutler} {and} \bibinfo{person}{Leo
  Breiman}.} \bibinfo{year}{1994}\natexlab{}.
\newblock \showarticletitle{Archetypal analysis}.
\newblock \bibinfo{journal}{\emph{Technometrics}} \bibinfo{number}{36}
  (\bibinfo{year}{1994}), \bibinfo{pages}{338--347}.
\newblock
Issue 4.


\bibitem[\protect\citeauthoryear{Deng, Dong, Socher, Li, Li, and Fei-Fei}{Deng
  et~al\mbox{.}}{2009}]%
        {deng2009imagenet}
\bibfield{author}{\bibinfo{person}{Jia Deng}, \bibinfo{person}{Wei Dong},
  \bibinfo{person}{Richard Socher}, \bibinfo{person}{Li-Jia Li},
  \bibinfo{person}{Kai Li}, {and} \bibinfo{person}{Li Fei-Fei}.}
  \bibinfo{year}{2009}\natexlab{}.
\newblock \showarticletitle{Imagenet: A large-scale hierarchical image
  database}. In \bibinfo{booktitle}{\emph{2009 IEEE conference on computer
  vision and pattern recognition}}. Ieee, \bibinfo{pages}{248--255}.
\newblock


\bibitem[\protect\citeauthoryear{Ding, Hardt, Miller, and Schmidt}{Ding
  et~al\mbox{.}}{2021}]%
        {ding2021retiring}
\bibfield{author}{\bibinfo{person}{Frances Ding}, \bibinfo{person}{Moritz
  Hardt}, \bibinfo{person}{John Miller}, {and} \bibinfo{person}{Ludwig
  Schmidt}.} \bibinfo{year}{2021}\natexlab{}.
\newblock \showarticletitle{Retiring Adult: New Datasets for Fair Machine
  Learning}.
\newblock \bibinfo{journal}{\emph{Advances in Neural Information Processing
  Systems}}  \bibinfo{volume}{34} (\bibinfo{year}{2021}).
\newblock


\bibitem[\protect\citeauthoryear{dos Santos~Machado, Ballester, Cao, Mwangi,
  Caldieraro, Kapczinski, and Passos}{dos Santos~Machado et~al\mbox{.}}{2021}]%
        {Machado2020}
\bibfield{author}{\bibinfo{person}{Cristiane dos Santos~Machado},
  \bibinfo{person}{Pedro~L. Ballester}, \bibinfo{person}{Bo Cao},
  \bibinfo{person}{Benson Mwangi}, \bibinfo{person}{Marco~Antonio Caldieraro},
  \bibinfo{person}{Flávio Kapczinski}, {and} \bibinfo{person}{Ives~Cavalcante
  Passos}.} \bibinfo{year}{2021}\natexlab{}.
\newblock \showarticletitle{Prediction of suicide attempts in a prospective
  cohort study with a nationally representative sample of the US population}.
\newblock \bibinfo{journal}{\emph{Psychological Medicine}}
  (\bibinfo{year}{2021}), \bibinfo{pages}{1--12}.
\newblock


\bibitem[\protect\citeauthoryear{Dotan and Milli}{Dotan and Milli}{2020}]%
        {DotanMilli2020}
\bibfield{author}{\bibinfo{person}{Ravit Dotan} {and} \bibinfo{person}{Smitha
  Milli}.} \bibinfo{year}{2020}\natexlab{}.
\newblock \showarticletitle{Value-laden Disciplinary Shifts in Machine
  Learning}.
\newblock \bibinfo{journal}{\emph{FAT* '20, January 27-30, 2020, Barcelona,
  Spain}} (\bibinfo{year}{2020}).
\newblock


\bibitem[\protect\citeauthoryear{Dwork, Hardt, Pitassi, Reingold, and
  Zemel}{Dwork et~al\mbox{.}}{2012}]%
        {dwork2012fairness}
\bibfield{author}{\bibinfo{person}{Cynthia Dwork}, \bibinfo{person}{Moritz
  Hardt}, \bibinfo{person}{Toniann Pitassi}, \bibinfo{person}{Omer Reingold},
  {and} \bibinfo{person}{Richard Zemel}.} \bibinfo{year}{2012}\natexlab{}.
\newblock \showarticletitle{Fairness through awareness}. In
  \bibinfo{booktitle}{\emph{Proceedings of the 3rd innovations in theoretical
  computer science conference}}. \bibinfo{pages}{214--226}.
\newblock


\bibitem[\protect\citeauthoryear{D’Excelle}{D’Excelle}{2014}]%
        {D'Excelle}
\bibfield{author}{\bibinfo{person}{Ben D’Excelle}.}
  \bibinfo{year}{2014}\natexlab{}.
\newblock \bibinfo{booktitle}{\emph{Representative Sample}}.
\newblock \bibinfo{publisher}{Springer}. 5511--5513 pages.
\newblock
\urldef\tempurl%
\url{https://doi.org/10.1007/978-94-007-0753-5_2476}
\showDOI{\tempurl}


\bibitem[\protect\citeauthoryear{Fisher}{Fisher}{1935}]%
        {Fisher1935}
\bibfield{author}{\bibinfo{person}{Ronald~A. Fisher}.}
  \bibinfo{year}{1935}\natexlab{}.
\newblock \bibinfo{booktitle}{\emph{The Design of Experiments}}.
\newblock \bibinfo{publisher}{Oliver and Boyd}.
\newblock


\bibitem[\protect\citeauthoryear{Gautier, Polito, Bardenet, and Valko}{Gautier
  et~al\mbox{.}}{2019}]%
        {gautier2019dppy}
\bibfield{author}{\bibinfo{person}{Guillaume Gautier},
  \bibinfo{person}{Guillermo Polito}, \bibinfo{person}{R{\'e}mi Bardenet},
  {and} \bibinfo{person}{Michal Valko}.} \bibinfo{year}{2019}\natexlab{}.
\newblock \showarticletitle{DPPy: DPP Sampling with Python.}
\newblock \bibinfo{journal}{\emph{J. Mach. Learn. Res.}}  \bibinfo{volume}{20}
  (\bibinfo{year}{2019}), \bibinfo{pages}{180--1}.
\newblock


\bibitem[\protect\citeauthoryear{Gebru, Morgenstern, Vechhione, Vaughan,
  Wallach, III, and Crawford}{Gebru et~al\mbox{.}}{2021}]%
        {Gebru2021}
\bibfield{author}{\bibinfo{person}{Timnit Gebru}, \bibinfo{person}{Jamie
  Morgenstern}, \bibinfo{person}{Briana Vechhione},
  \bibinfo{person}{Jennifer~Wrotmen Vaughan}, \bibinfo{person}{Hanna Wallach},
  \bibinfo{person}{Hal~Daume III}, {and} \bibinfo{person}{Kate Crawford}.}
  \bibinfo{year}{2021}\natexlab{}.
\newblock \showarticletitle{Datasheets for Datasets}.
\newblock \bibinfo{journal}{\emph{arXiv:1803.09010v8}} (\bibinfo{year}{2021}).
\newblock


\bibitem[\protect\citeauthoryear{Ghojogh, Nekoei, Ghojogh, Karray, and
  Crowley}{Ghojogh et~al\mbox{.}}{2020}]%
        {ghojogh2020}
\bibfield{author}{\bibinfo{person}{Benyamin Ghojogh}, \bibinfo{person}{Hadi
  Nekoei}, \bibinfo{person}{Aydin Ghojogh}, \bibinfo{person}{Fakhri Karray},
  {and} \bibinfo{person}{Mark Crowley}.} \bibinfo{year}{2020}\natexlab{}.
\newblock \showarticletitle{Sampling algorithms, from Survey Sampling to Monte
  Carlo Methods: Tutorial and Literature Review}.
\newblock \bibinfo{journal}{\emph{arXiv:2011.00901v1}} (\bibinfo{year}{2020}).
\newblock


\bibitem[\protect\citeauthoryear{Gideon}{Gideon}{2012}]%
        {Gideon2012}
\bibfield{author}{\bibinfo{person}{Lior Gideon}.}
  \bibinfo{year}{2012}\natexlab{}.
\newblock \bibinfo{booktitle}{\emph{Handbook of Survey Methodology for the
  Social Sciences}}.
\newblock \bibinfo{publisher}{Springer}.
\newblock
\showISBNx{ISBN: 978-1-4614-3876-2}


\bibitem[\protect\citeauthoryear{Gilpin}{Gilpin}{2021}]%
        {Survey_106}
\bibfield{author}{\bibinfo{person}{William Gilpin}.}
  \bibinfo{year}{2021}\natexlab{}.
\newblock \showarticletitle{Chaos as an interpretable benchmark for forecasting
  and data-driven modelling}.
\newblock \bibinfo{journal}{\emph{arXiv preprint arXiv:2110.05266}}
  (\bibinfo{year}{2021}).
\newblock


\bibitem[\protect\citeauthoryear{Gong, Chao, Grauman, and Sha}{Gong
  et~al\mbox{.}}{2014}]%
        {gong2014diverse}
\bibfield{author}{\bibinfo{person}{Boqing Gong}, \bibinfo{person}{Wei-Lun
  Chao}, \bibinfo{person}{Kristen Grauman}, {and} \bibinfo{person}{Fei Sha}.}
  \bibinfo{year}{2014}\natexlab{}.
\newblock \showarticletitle{Diverse sequential subset selection for supervised
  video summarization}.
\newblock \bibinfo{journal}{\emph{Advances in neural information processing
  systems}}  \bibinfo{volume}{27} (\bibinfo{year}{2014}).
\newblock


\bibitem[\protect\citeauthoryear{Gretton, Borgwardt, Rasch, Sch{\"o}lkopf, and
  Smola}{Gretton et~al\mbox{.}}{2012}]%
        {gretton2012kernelMMD}
\bibfield{author}{\bibinfo{person}{Arthur Gretton}, \bibinfo{person}{Karsten~M
  Borgwardt}, \bibinfo{person}{Malte~J Rasch}, \bibinfo{person}{Bernhard
  Sch{\"o}lkopf}, {and} \bibinfo{person}{Alexander Smola}.}
  \bibinfo{year}{2012}\natexlab{}.
\newblock \showarticletitle{A kernel two-sample test}.
\newblock \bibinfo{journal}{\emph{The Journal of Machine Learning Research}}
  \bibinfo{volume}{13}, \bibinfo{number}{1} (\bibinfo{year}{2012}),
  \bibinfo{pages}{723--773}.
\newblock


\bibitem[\protect\citeauthoryear{Gy}{Gy}{1998}]%
        {Gy1998}
\bibfield{author}{\bibinfo{person}{Pierre Gy}.}
  \bibinfo{year}{1998}\natexlab{}.
\newblock \bibinfo{booktitle}{\emph{Sampling for Analytical Purposes}}.
\newblock \bibinfo{publisher}{Wiley}.
\newblock
\showISBNx{ISBN: 978-0-471-97956-2}


\bibitem[\protect\citeauthoryear{Hardt, Price, and Srebro}{Hardt
  et~al\mbox{.}}{2016}]%
        {hardt2016equality}
\bibfield{author}{\bibinfo{person}{Moritz Hardt}, \bibinfo{person}{Eric Price},
  {and} \bibinfo{person}{Nati Srebro}.} \bibinfo{year}{2016}\natexlab{}.
\newblock \showarticletitle{Equality of opportunity in supervised learning}.
\newblock \bibinfo{journal}{\emph{Advances in neural information processing
  systems}}  \bibinfo{volume}{29} (\bibinfo{year}{2016}),
  \bibinfo{pages}{3315--3323}.
\newblock


\bibitem[\protect\citeauthoryear{Hibberts, Johnson, and Hudson}{Hibberts
  et~al\mbox{.}}{2012}]%
        {Gideon2012-5}
\bibfield{author}{\bibinfo{person}{Mary Hibberts}, \bibinfo{person}{R.~Burke
  Johnson}, {and} \bibinfo{person}{Kenneth Hudson}.}
  \bibinfo{year}{2012}\natexlab{}.
\newblock \bibinfo{booktitle}{\emph{Common Survey Sampling Techniques}}.
\newblock \bibinfo{publisher}{Springer}. 53--74 pages.
\newblock
\showISBNx{ISBN: 978-1-4614-3876-2}


\bibitem[\protect\citeauthoryear{Hou, Li, Li, and Liu}{Hou
  et~al\mbox{.}}{2019}]%
        {hou2019density}
\bibfield{author}{\bibinfo{person}{Yun Hou}, \bibinfo{person}{Bailin Li},
  \bibinfo{person}{Li Li}, {and} \bibinfo{person}{Jiajia Liu}.}
  \bibinfo{year}{2019}\natexlab{}.
\newblock \showarticletitle{A density-based under-sampling algorithm for
  imbalance classification}. In \bibinfo{booktitle}{\emph{Journal of Physics:
  Conference Series}}, Vol.~\bibinfo{volume}{1302}. IOP Publishing,
  \bibinfo{pages}{022064}.
\newblock


\bibitem[\protect\citeauthoryear{Huang}{Huang}{2021}]%
        {Huang2021}
\bibfield{author}{\bibinfo{person}{Jonathan~Yinhao Huang}.}
  \bibinfo{year}{2021}\natexlab{}.
\newblock \showarticletitle{Representativeness Is Not Representative -
  Addressing Major Inferential Threats in the UK Biobank and Other Big Data
  Repositories}.
\newblock \bibinfo{journal}{\emph{Epidemiology}} \bibinfo{number}{32}
  (\bibinfo{year}{2021}), \bibinfo{pages}{189--193}.
\newblock
Issue 2.


\bibitem[\protect\citeauthoryear{Huang, Wang, Blaney, Slaughter, McKeon, Zhou,
  Jacob, and Hughes}{Huang et~al\mbox{.}}{2021}]%
        {Survey_71}
\bibfield{author}{\bibinfo{person}{Zhe Huang}, \bibinfo{person}{Liang Wang},
  \bibinfo{person}{Giles Blaney}, \bibinfo{person}{Christopher Slaughter},
  \bibinfo{person}{Devon McKeon}, \bibinfo{person}{Ziyu Zhou},
  \bibinfo{person}{Robert Jacob}, {and} \bibinfo{person}{Michael~C Hughes}.}
  \bibinfo{year}{2021}\natexlab{}.
\newblock \showarticletitle{The Tufts fNIRS Mental Workload Dataset \&
  Benchmark for Brain-Computer Interfaces that Generalize}.
\newblock  (\bibinfo{year}{2021}).
\newblock


\bibitem[\protect\citeauthoryear{Jin, Xu, Sun, Asudeh, and Jagadish}{Jin
  et~al\mbox{.}}{2020}]%
        {jin2020mithracoverage}
\bibfield{author}{\bibinfo{person}{Zhongjun Jin}, \bibinfo{person}{Mengjing
  Xu}, \bibinfo{person}{Chenkai Sun}, \bibinfo{person}{Abolfazl Asudeh}, {and}
  \bibinfo{person}{HV Jagadish}.} \bibinfo{year}{2020}\natexlab{}.
\newblock \showarticletitle{Mithracoverage: a system for investigating
  population bias for intersectional fairness}. In
  \bibinfo{booktitle}{\emph{Proceedings of the 2020 ACM SIGMOD International
  Conference on Management of Data}}. \bibinfo{pages}{2721--2724}.
\newblock


\bibitem[\protect\citeauthoryear{Kalsbeek}{Kalsbeek}{2003}]%
        {kalsbeek2003sampling}
\bibfield{author}{\bibinfo{person}{William~D Kalsbeek}.}
  \bibinfo{year}{2003}\natexlab{}.
\newblock \showarticletitle{Sampling minority groups in health surveys}.
\newblock \bibinfo{journal}{\emph{Statistics in Medicine}}
  \bibinfo{volume}{22}, \bibinfo{number}{9} (\bibinfo{year}{2003}),
  \bibinfo{pages}{1527--1549}.
\newblock


\bibitem[\protect\citeauthoryear{Kalsbeek, Boyle, Agans, and White}{Kalsbeek
  et~al\mbox{.}}{2007}]%
        {kalsbeek2007disproportionate}
\bibfield{author}{\bibinfo{person}{William~D Kalsbeek},
  \bibinfo{person}{Walter~R Boyle}, \bibinfo{person}{Robert~P Agans}, {and}
  \bibinfo{person}{John~E White}.} \bibinfo{year}{2007}\natexlab{}.
\newblock \showarticletitle{Disproportionate sampling for population subgroups
  in telephone surveys}.
\newblock \bibinfo{journal}{\emph{Statistics in medicine}}
  \bibinfo{volume}{26}, \bibinfo{number}{8} (\bibinfo{year}{2007}),
  \bibinfo{pages}{1657--1674}.
\newblock


\bibitem[\protect\citeauthoryear{Kantorovich}{Kantorovich}{1960}]%
        {kantorovich1960mathematicalEMD2}
\bibfield{author}{\bibinfo{person}{Leonid~V Kantorovich}.}
  \bibinfo{year}{1960}\natexlab{}.
\newblock \showarticletitle{Mathematical methods of organizing and planning
  production}.
\newblock \bibinfo{journal}{\emph{Management science}} \bibinfo{volume}{6},
  \bibinfo{number}{4} (\bibinfo{year}{1960}), \bibinfo{pages}{366--422}.
\newblock


\bibitem[\protect\citeauthoryear{Kaushal, Altman, and Langlotz}{Kaushal
  et~al\mbox{.}}{2020}]%
        {kaushal2020geographic}
\bibfield{author}{\bibinfo{person}{Amit Kaushal}, \bibinfo{person}{Russ
  Altman}, {and} \bibinfo{person}{Curt Langlotz}.}
  \bibinfo{year}{2020}\natexlab{}.
\newblock \showarticletitle{Geographic distribution of US cohorts used to train
  deep learning algorithms}.
\newblock \bibinfo{journal}{\emph{Jama}} \bibinfo{volume}{324},
  \bibinfo{number}{12} (\bibinfo{year}{2020}), \bibinfo{pages}{1212--1213}.
\newblock


\bibitem[\protect\citeauthoryear{Kelly, Karthikesalingam, Suleyman, Corrado,
  and King}{Kelly et~al\mbox{.}}{2019}]%
        {Kelly2019}
\bibfield{author}{\bibinfo{person}{Christopher~J. Kelly}, \bibinfo{person}{Alan
  Karthikesalingam}, \bibinfo{person}{Mustafa Suleyman}, \bibinfo{person}{Greg
  Corrado}, {and} \bibinfo{person}{Dominic King}.}
  \bibinfo{year}{2019}\natexlab{}.
\newblock \showarticletitle{Key challenges for delivering clinical impact with
  artificial intelligence}.
\newblock \bibinfo{journal}{\emph{BMC Medicine}} \bibinfo{number}{17}
  (\bibinfo{year}{2019}).
\newblock
Issue 195.


\bibitem[\protect\citeauthoryear{Kiskin, Sinka, Cobb, Rafique, Wang, Zilli,
  Gutteridge, Dam, Marinos, Li, et~al\mbox{.}}{Kiskin et~al\mbox{.}}{2021}]%
        {Survey_44}
\bibfield{author}{\bibinfo{person}{Ivan Kiskin}, \bibinfo{person}{Marianne
  Sinka}, \bibinfo{person}{Adam~D Cobb}, \bibinfo{person}{Waqas Rafique},
  \bibinfo{person}{Lawrence Wang}, \bibinfo{person}{Davide Zilli},
  \bibinfo{person}{Benjamin Gutteridge}, \bibinfo{person}{Rinita Dam},
  \bibinfo{person}{Theodoros Marinos}, \bibinfo{person}{Yunpeng Li},
  {et~al\mbox{.}}} \bibinfo{year}{2021}\natexlab{}.
\newblock \showarticletitle{HumBugDB: a large-scale acoustic mosquito dataset}.
\newblock \bibinfo{journal}{\emph{arXiv preprint arXiv:2110.07607}}
  (\bibinfo{year}{2021}).
\newblock


\bibitem[\protect\citeauthoryear{Kj{\ae}rsgaard, Gr{\o}nberg, and
  Clemmensen}{Kj{\ae}rsgaard et~al\mbox{.}}{2021}]%
        {kjaersgaard2021sampling}
\bibfield{author}{\bibinfo{person}{Rune~D Kj{\ae}rsgaard},
  \bibinfo{person}{Manja~G Gr{\o}nberg}, {and} \bibinfo{person}{Line~KH
  Clemmensen}.} \bibinfo{year}{2021}\natexlab{}.
\newblock \showarticletitle{Sampling To Improve Predictions For
  Underrepresented Observations In Imbalanced Data}.
\newblock \bibinfo{journal}{\emph{arXiv preprint arXiv:2111.09065}}
  (\bibinfo{year}{2021}).
\newblock


\bibitem[\protect\citeauthoryear{Kohavi and Becker}{Kohavi and Becker}{1996}]%
        {kohavi1996adult}
\bibfield{author}{\bibinfo{person}{Ronny Kohavi} {and} \bibinfo{person}{Barry
  Becker}.} \bibinfo{year}{1996}\natexlab{}.
\newblock \showarticletitle{Adult data set}.
\newblock \bibinfo{journal}{\emph{UCI machine learning repository}}
  \bibinfo{volume}{5} (\bibinfo{year}{1996}), \bibinfo{pages}{2093}.
\newblock


\bibitem[\protect\citeauthoryear{Kondmann, Toker, Ru{\ss}wurm, Camero~Unzueta,
  Peressuti, Milcinski, Long{\'e}p{\'e}, Mathieu, Davis, Marchisio,
  et~al\mbox{.}}{Kondmann et~al\mbox{.}}{2021}]%
        {Survey_36}
\bibfield{author}{\bibinfo{person}{Lukas Kondmann}, \bibinfo{person}{Aysim
  Toker}, \bibinfo{person}{Marc Ru{\ss}wurm}, \bibinfo{person}{Andres
  Camero~Unzueta}, \bibinfo{person}{Devis Peressuti}, \bibinfo{person}{Grega
  Milcinski}, \bibinfo{person}{Nicolas Long{\'e}p{\'e}},
  \bibinfo{person}{Pierre-Philippe Mathieu}, \bibinfo{person}{Timothy Davis},
  \bibinfo{person}{Giovanni Marchisio}, {et~al\mbox{.}}}
  \bibinfo{year}{2021}\natexlab{}.
\newblock \showarticletitle{DENETHOR: The DynamicEarthNET dataset for
  Harmonized, inter-Operable, analysis-Ready, daily crop monitoring from
  space}. In \bibinfo{booktitle}{\emph{Thirty-fifth Conference on Neural
  Information Processing Systems Datasets and Benchmarks Track}}.
\newblock


\bibitem[\protect\citeauthoryear{Korosteleva and Lee}{Korosteleva and
  Lee}{2021}]%
        {Survey_1}
\bibfield{author}{\bibinfo{person}{Maria Korosteleva} {and}
  \bibinfo{person}{Sung-Hee Lee}.} \bibinfo{year}{2021}\natexlab{}.
\newblock \showarticletitle{Generating Datasets of 3D Garments with Sewing
  Patterns}.
\newblock \bibinfo{journal}{\emph{arXiv preprint arXiv:2109.05633}}
  (\bibinfo{year}{2021}).
\newblock


\bibitem[\protect\citeauthoryear{Krause, Singh, and Guestrin}{Krause
  et~al\mbox{.}}{2008}]%
        {krause2008near}
\bibfield{author}{\bibinfo{person}{Andreas Krause}, \bibinfo{person}{Ajit
  Singh}, {and} \bibinfo{person}{Carlos Guestrin}.}
  \bibinfo{year}{2008}\natexlab{}.
\newblock \showarticletitle{Near-optimal sensor placements in Gaussian
  processes: Theory, efficient algorithms and empirical studies.}
\newblock \bibinfo{journal}{\emph{Journal of Machine Learning Research}}
  \bibinfo{volume}{9}, \bibinfo{number}{2} (\bibinfo{year}{2008}).
\newblock


\bibitem[\protect\citeauthoryear{Kruskal and Mosteller}{Kruskal and
  Mosteller}{1979a}]%
        {KruskalMosteller1979-1}
\bibfield{author}{\bibinfo{person}{William Kruskal} {and}
  \bibinfo{person}{Frederick Mosteller}.} \bibinfo{year}{1979}\natexlab{a}.
\newblock \showarticletitle{Representative sampling, I: Non-scientific
  Literature}.
\newblock \bibinfo{journal}{\emph{International Statistical Review}}
  \bibinfo{number}{47} (\bibinfo{year}{1979}), \bibinfo{pages}{13--24}.
\newblock


\bibitem[\protect\citeauthoryear{Kruskal and Mosteller}{Kruskal and
  Mosteller}{1979b}]%
        {KruskalMosteller1979-2}
\bibfield{author}{\bibinfo{person}{William Kruskal} {and}
  \bibinfo{person}{Frederick Mosteller}.} \bibinfo{year}{1979}\natexlab{b}.
\newblock \showarticletitle{Representative sampling, II: Scientific Literature,
  Excluding Statistics}.
\newblock \bibinfo{journal}{\emph{International Statistical Review}}
  \bibinfo{number}{47} (\bibinfo{year}{1979}), \bibinfo{pages}{111--127}.
\newblock


\bibitem[\protect\citeauthoryear{Kruskal and Mosteller}{Kruskal and
  Mosteller}{1979c}]%
        {KruskalMosteller1979-3}
\bibfield{author}{\bibinfo{person}{William Kruskal} {and}
  \bibinfo{person}{Frederick Mosteller}.} \bibinfo{year}{1979}\natexlab{c}.
\newblock \showarticletitle{Representative sampling, III: the Current
  Statistical Literature}.
\newblock \bibinfo{journal}{\emph{International Statistical Review}}
  \bibinfo{number}{47} (\bibinfo{year}{1979}), \bibinfo{pages}{245--265}.
\newblock


\bibitem[\protect\citeauthoryear{Kruskal and Mosteller}{Kruskal and
  Mosteller}{1980}]%
        {KruskalMosteller1979-4}
\bibfield{author}{\bibinfo{person}{William Kruskal} {and}
  \bibinfo{person}{Frederick Mosteller}.} \bibinfo{year}{1980}\natexlab{}.
\newblock \showarticletitle{Representative sampling, IV: the History of the
  Concept in Statics, 1895-1939}.
\newblock \bibinfo{journal}{\emph{International Statistical Review}}
  \bibinfo{number}{48} (\bibinfo{year}{1980}), \bibinfo{pages}{169--195}.
\newblock


\bibitem[\protect\citeauthoryear{Kulahci, Frumosu, Khan, Ørnskov Rønsch, and
  Spooner}{Kulahci et~al\mbox{.}}{2020}]%
        {Kulahci2020}
\bibfield{author}{\bibinfo{person}{Murat Kulahci},
  \bibinfo{person}{Flavia~Dalia Frumosu}, \bibinfo{person}{Abdul~Rauf Khan},
  \bibinfo{person}{Georg Ørnskov Rønsch}, {and} \bibinfo{person}{Max~Peter
  Spooner}.} \bibinfo{year}{2020}\natexlab{}.
\newblock \showarticletitle{Experiences with big data: Accounts from a data
  scientist’s perspective}.
\newblock \bibinfo{journal}{\emph{Quality Engineering}} \bibinfo{number}{32}
  (\bibinfo{year}{2020}), \bibinfo{pages}{529--542}.
\newblock
Issue 4.


\bibitem[\protect\citeauthoryear{Kulesza and Taskar}{Kulesza and
  Taskar}{2011}]%
        {kulesza2011k-dpp}
\bibfield{author}{\bibinfo{person}{Alex Kulesza} {and} \bibinfo{person}{Ben
  Taskar}.} \bibinfo{year}{2011}\natexlab{}.
\newblock \showarticletitle{k-DPPs: Fixed-size determinantal point processes}.
  In \bibinfo{booktitle}{\emph{ICML}}.
\newblock


\bibitem[\protect\citeauthoryear{Kulesza, Taskar, et~al\mbox{.}}{Kulesza
  et~al\mbox{.}}{2012a}]%
        {kulesza2012determinantal}
\bibfield{author}{\bibinfo{person}{Alex Kulesza}, \bibinfo{person}{Ben Taskar},
  {et~al\mbox{.}}} \bibinfo{year}{2012}\natexlab{a}.
\newblock \showarticletitle{Determinantal point processes for machine
  learning}.
\newblock \bibinfo{journal}{\emph{Foundations and Trends{\textregistered} in
  Machine Learning}} \bibinfo{volume}{5}, \bibinfo{number}{2--3}
  (\bibinfo{year}{2012}), \bibinfo{pages}{123--286}.
\newblock


\bibitem[\protect\citeauthoryear{Kulesza, Taskar, et~al\mbox{.}}{Kulesza
  et~al\mbox{.}}{2012b}]%
        {DPPforML}
\bibfield{author}{\bibinfo{person}{Alex Kulesza}, \bibinfo{person}{Ben Taskar},
  {et~al\mbox{.}}} \bibinfo{year}{2012}\natexlab{b}.
\newblock \showarticletitle{Determinantal point processes for machine
  learning}.
\newblock \bibinfo{journal}{\emph{Foundations and Trends{\textregistered} in
  Machine Learning}} \bibinfo{volume}{5}, \bibinfo{number}{2--3}
  (\bibinfo{year}{2012}), \bibinfo{pages}{123--286}.
\newblock


\bibitem[\protect\citeauthoryear{Lee, Taddy, and Gray}{Lee
  et~al\mbox{.}}{2010}]%
        {Lee2010}
\bibfield{author}{\bibinfo{person}{Herbert K.~H. Lee}, \bibinfo{person}{Matthew
  Taddy}, {and} \bibinfo{person}{Genetha~A. Gray}.}
  \bibinfo{year}{2010}\natexlab{}.
\newblock \showarticletitle{Selection of a Representative Sample}.
\newblock \bibinfo{journal}{\emph{Journal of Classification}}
  \bibinfo{number}{27} (\bibinfo{year}{2010}), \bibinfo{pages}{41--53}.
\newblock


\bibitem[\protect\citeauthoryear{Lehmann, Romano, and Casella}{Lehmann
  et~al\mbox{.}}{2005}]%
        {lehmann2005testing}
\bibfield{author}{\bibinfo{person}{Erich~Leo Lehmann},
  \bibinfo{person}{Joseph~P Romano}, {and} \bibinfo{person}{George Casella}.}
  \bibinfo{year}{2005}\natexlab{}.
\newblock \bibinfo{booktitle}{\emph{Testing statistical hypotheses}}.
  Vol.~\bibinfo{volume}{3}.
\newblock \bibinfo{publisher}{Springer}.
\newblock


\bibitem[\protect\citeauthoryear{Li, Guo, Ren, Cong, Hou, Kwong, and Tao}{Li
  et~al\mbox{.}}{2020}]%
        {Li2019}
\bibfield{author}{\bibinfo{person}{Chongyi Li}, \bibinfo{person}{Chunle Guo},
  \bibinfo{person}{Wenqi Ren}, \bibinfo{person}{Runmin Cong},
  \bibinfo{person}{Junhui Hou}, \bibinfo{person}{Sam Kwong}, {and}
  \bibinfo{person}{Dacheng Tao}.} \bibinfo{year}{2020}\natexlab{}.
\newblock \showarticletitle{An Underwater Image Enhancement Benchmark Dataset
  and Beyond}.
\newblock \bibinfo{journal}{\emph{IEEE Transactions on Image Processing}}
  \bibinfo{number}{29} (\bibinfo{year}{2020}), \bibinfo{pages}{4376--4389}.
\newblock
Issue 1.


\bibitem[\protect\citeauthoryear{Lin and Bilmes}{Lin and Bilmes}{2012}]%
        {lin2012learning}
\bibfield{author}{\bibinfo{person}{Hui Lin} {and} \bibinfo{person}{Jeff~A
  Bilmes}.} \bibinfo{year}{2012}\natexlab{}.
\newblock \showarticletitle{Learning mixtures of submodular shells with
  application to document summarization}.
\newblock \bibinfo{journal}{\emph{arXiv preprint arXiv:1210.4871}}
  (\bibinfo{year}{2012}).
\newblock


\bibitem[\protect\citeauthoryear{Liu, Khandagale, White, and Neiswanger}{Liu
  et~al\mbox{.}}{2021}]%
        {Survey_74}
\bibfield{author}{\bibinfo{person}{Yang Liu}, \bibinfo{person}{Sujay
  Khandagale}, \bibinfo{person}{Colin White}, {and} \bibinfo{person}{Willie
  Neiswanger}.} \bibinfo{year}{2021}\natexlab{}.
\newblock \showarticletitle{Synthetic benchmarks for scientific research in
  explainable machine learning}.
\newblock \bibinfo{journal}{\emph{arXiv preprint arXiv:2106.12543}}
  (\bibinfo{year}{2021}).
\newblock


\bibitem[\protect\citeauthoryear{Log{\'e}, Ross, Dadey, Jain, Saporta, Ng, and
  Rajpurkar}{Log{\'e} et~al\mbox{.}}{2021}]%
        {Survey_43}
\bibfield{author}{\bibinfo{person}{C{\'e}cile Log{\'e}}, \bibinfo{person}{Emily
  Ross}, \bibinfo{person}{David Yaw~Amoah Dadey}, \bibinfo{person}{Saahil
  Jain}, \bibinfo{person}{Adriel Saporta}, \bibinfo{person}{Andrew~Y Ng}, {and}
  \bibinfo{person}{Pranav Rajpurkar}.} \bibinfo{year}{2021}\natexlab{}.
\newblock \showarticletitle{Q-Pain: A Question Answering Dataset to Measure
  Social Bias in Pain Management}.
\newblock \bibinfo{journal}{\emph{arXiv preprint arXiv:2108.01764}}
  (\bibinfo{year}{2021}).
\newblock


\bibitem[\protect\citeauthoryear{Lowry and G.~Macpherson}{Lowry and
  G.~Macpherson}{1988}]%
        {Lowry}
\bibfield{author}{\bibinfo{person}{S. Lowry} {and} \bibinfo{person}{“A blot
  on the~profession G.~Macpherson}.} \bibinfo{year}{1988}\natexlab{}.
\newblock \bibinfo{journal}{\emph{British Medical Journal}}
  \bibinfo{volume}{296}, \bibinfo{number}{6623} (\bibinfo{year}{1988}),
  \bibinfo{pages}{657--658}.
\newblock


\bibitem[\protect\citeauthoryear{Lum and Isaac}{Lum and Isaac}{2016}]%
        {lum2016predict}
\bibfield{author}{\bibinfo{person}{Kristian Lum} {and} \bibinfo{person}{William
  Isaac}.} \bibinfo{year}{2016}\natexlab{}.
\newblock \showarticletitle{To predict and serve?}
\newblock \bibinfo{journal}{\emph{Significance}} \bibinfo{volume}{13},
  \bibinfo{number}{5} (\bibinfo{year}{2016}), \bibinfo{pages}{14--19}.
\newblock


\bibitem[\protect\citeauthoryear{M.~Ali and Rieke}{M.~Ali and Rieke}{2019}]%
        {FacebookAd}
\bibfield{author}{\bibinfo{person}{M.~Bogen A. Korlova A.~Mislove M.~Ali,
  P.~Sapienzynski} {and} \bibinfo{person}{A. Rieke}.}
  \bibinfo{year}{2019}\natexlab{}.
\newblock \showarticletitle{Discrimination through Optimization: How
  Facebook’s Ad Delivery Can Lead to Biased Outcomes}.
\newblock \bibinfo{journal}{\emph{ACM on Human-Computer Interaction}}
  (\bibinfo{year}{2019}).
\newblock


\bibitem[\protect\citeauthoryear{MacKay}{MacKay}{2005}]%
        {MacKay2003}
\bibfield{author}{\bibinfo{person}{David~J.C. MacKay}.}
  \bibinfo{year}{2005}\natexlab{}.
\newblock \bibinfo{booktitle}{\emph{Informaiton Theory, Inference, and Learning
  Algorithms} (\bibinfo{edition}{7} ed.)}.
\newblock \bibinfo{publisher}{Cambridge University Press}.
\newblock
\showISBNx{ISBN: 978-1-4614-3876-2}


\bibitem[\protect\citeauthoryear{Malinin, Band, Chesnokov, Gal, Gales, Noskov,
  Ploskonosov, Prokhorenkova, Provilkov, Raina, et~al\mbox{.}}{Malinin
  et~al\mbox{.}}{2021}]%
        {Survey_69}
\bibfield{author}{\bibinfo{person}{Andrey Malinin}, \bibinfo{person}{Neil
  Band}, \bibinfo{person}{German Chesnokov}, \bibinfo{person}{Yarin Gal},
  \bibinfo{person}{Mark~JF Gales}, \bibinfo{person}{Alexey Noskov},
  \bibinfo{person}{Andrey Ploskonosov}, \bibinfo{person}{Liudmila
  Prokhorenkova}, \bibinfo{person}{Ivan Provilkov}, \bibinfo{person}{Vatsal
  Raina}, {et~al\mbox{.}}} \bibinfo{year}{2021}\natexlab{}.
\newblock \showarticletitle{Shifts: A dataset of real distributional shift
  across multiple large-scale tasks}.
\newblock \bibinfo{journal}{\emph{arXiv preprint arXiv:2107.07455}}
  (\bibinfo{year}{2021}).
\newblock


\bibitem[\protect\citeauthoryear{Mao, Niu, Jiang, Liang, Chen, Liang, Li, Ye,
  Zhang, Li, et~al\mbox{.}}{Mao et~al\mbox{.}}{2021}]%
        {Survey_47}
\bibfield{author}{\bibinfo{person}{Jiageng Mao}, \bibinfo{person}{Minzhe Niu},
  \bibinfo{person}{Chenhan Jiang}, \bibinfo{person}{Hanxue Liang},
  \bibinfo{person}{Jingheng Chen}, \bibinfo{person}{Xiaodan Liang},
  \bibinfo{person}{Yamin Li}, \bibinfo{person}{Chaoqiang Ye},
  \bibinfo{person}{Wei Zhang}, \bibinfo{person}{Zhenguo Li}, {et~al\mbox{.}}}
  \bibinfo{year}{2021}\natexlab{}.
\newblock \showarticletitle{One million scenes for autonomous driving: Once
  dataset}.
\newblock \bibinfo{journal}{\emph{arXiv preprint arXiv:2106.11037}}
  (\bibinfo{year}{2021}).
\newblock


\bibitem[\protect\citeauthoryear{Mazumder, Chitlangia, Banbury, Kang, Ciro,
  Achorn, Galvez, Sabini, Mattson, Kanter, et~al\mbox{.}}{Mazumder
  et~al\mbox{.}}{2021}]%
        {Survey_108}
\bibfield{author}{\bibinfo{person}{Mark Mazumder}, \bibinfo{person}{Sharad
  Chitlangia}, \bibinfo{person}{Colby Banbury}, \bibinfo{person}{Yiping Kang},
  \bibinfo{person}{Juan~Manuel Ciro}, \bibinfo{person}{Keith Achorn},
  \bibinfo{person}{Daniel Galvez}, \bibinfo{person}{Mark Sabini},
  \bibinfo{person}{Peter Mattson}, \bibinfo{person}{David Kanter},
  {et~al\mbox{.}}} \bibinfo{year}{2021}\natexlab{}.
\newblock \showarticletitle{Multilingual Spoken Words Corpus}. In
  \bibinfo{booktitle}{\emph{Thirty-fifth Conference on Neural Information
  Processing Systems Datasets and Benchmarks Track (Round 2)}}.
\newblock


\bibitem[\protect\citeauthoryear{Mehrabi, Morstatter, Saxena, Lerman, and
  Galstyan}{Mehrabi et~al\mbox{.}}{2021}]%
        {Mehrabi2019}
\bibfield{author}{\bibinfo{person}{Ninahreh Mehrabi}, \bibinfo{person}{Fred
  Morstatter}, \bibinfo{person}{Nripsuta Saxena}, \bibinfo{person}{Kristina
  Lerman}, {and} \bibinfo{person}{Aram Galstyan}.}
  \bibinfo{year}{2021}\natexlab{}.
\newblock \showarticletitle{A Survey on Bias and Fairness in Machine Learning}.
\newblock \bibinfo{journal}{\emph{Comput. Surveys}} \bibinfo{number}{54}
  (\bibinfo{year}{2021}).
\newblock
Issue 6.


\bibitem[\protect\citeauthoryear{MeriamWebster.com}{MeriamWebster.com}{2022}]%
        {MeriamWebster}
\bibfield{author}{\bibinfo{person}{MeriamWebster.com}.}
  \bibinfo{year}{2022}\natexlab{}.
\newblock \bibinfo{booktitle}{\emph{Definition of representative sampling}}.
\newblock
\urldef\tempurl%
\url{https://www.merriam-webster.com/dictionary/representative\%20sampling}
\showURL{%
\tempurl}


\bibitem[\protect\citeauthoryear{Mislove, S.~Lehmann, Ahn, p.~Onnela, and
  Rosenquist}{Mislove et~al\mbox{.}}{2011}]%
        {Mislove2011}
\bibfield{author}{\bibinfo{person}{A. Mislove}, \bibinfo{person}{S.
  S.~Lehmann}, \bibinfo{person}{Y.-Y. Ahn}, \bibinfo{person}{J. p. Onnela},
  {and} \bibinfo{person}{J. Rosenquist}.} \bibinfo{year}{2011}\natexlab{}.
\newblock \showarticletitle{Understanding the Demographics of Twitter Users}.
\newblock \bibinfo{journal}{\emph{Proceedings of: Fifth International AAAI
  Conference on Weblogs and Social Media}} \bibinfo{number}{5}
  (\bibinfo{year}{2011}), \bibinfo{pages}{554--557}.
\newblock
Issue 1.


\bibitem[\protect\citeauthoryear{Montgomery}{Montgomery}{2019}]%
        {Montgomery219}
\bibfield{author}{\bibinfo{person}{Douglas~C. Montgomery}.}
  \bibinfo{year}{2019}\natexlab{}.
\newblock \bibinfo{booktitle}{\emph{Design and Analysis of Experiments}
  (\bibinfo{edition}{10th} ed.)}.
\newblock \bibinfo{publisher}{Wiley}.
\newblock
\showISBNx{ISBN: 978-1-119-49244-3}


\bibitem[\protect\citeauthoryear{OECD}{OECD}{2022}]%
        {OECD}
\bibfield{author}{\bibinfo{person}{OECD}.} \bibinfo{year}{2022}\natexlab{}.
\newblock \bibinfo{booktitle}{\emph{May 26, 2022, Glossary of statistical
  terms, stats.oecd.org}}.
\newblock
\urldef\tempurl%
\url{https://stats.oecd.org/glossary/detail.asp?ID=3831}
\showURL{%
\tempurl}


\bibitem[\protect\citeauthoryear{Olteanu, CarlosCastillo, FernandoDiaz, and
  Kıcıman}{Olteanu et~al\mbox{.}}{2019}]%
        {Olteanu2019}
\bibfield{author}{\bibinfo{person}{Alexandra Olteanu},
  \bibinfo{person}{CarlosCastillo}, \bibinfo{person}{FernandoDiaz}, {and}
  \bibinfo{person}{Emre Kıcıman}.} \bibinfo{year}{2019}\natexlab{}.
\newblock \showarticletitle{Social Data: Biases, Methodological Pitfalls, and
  Ethical Boundaries}.
\newblock \bibinfo{journal}{\emph{Front. Big Data}} \bibinfo{number}{2}
  (\bibinfo{year}{2019}).
\newblock
Issue 13.


\bibitem[\protect\citeauthoryear{Otness, Gjoka, Bruna, Panozzo, Peherstorfer,
  Schneider, and Zorin}{Otness et~al\mbox{.}}{2021}]%
        {Survey_29}
\bibfield{author}{\bibinfo{person}{Karl Otness}, \bibinfo{person}{Arvi Gjoka},
  \bibinfo{person}{Joan Bruna}, \bibinfo{person}{Daniele Panozzo},
  \bibinfo{person}{Benjamin Peherstorfer}, \bibinfo{person}{Teseo Schneider},
  {and} \bibinfo{person}{Denis Zorin}.} \bibinfo{year}{2021}\natexlab{}.
\newblock \showarticletitle{An Extensible Benchmark Suite for Learning to
  Simulate Physical Systems}.
\newblock \bibinfo{journal}{\emph{arXiv preprint arXiv:2108.07799}}
  (\bibinfo{year}{2021}).
\newblock


\bibitem[\protect\citeauthoryear{Ou, Yuuan, and Cao}{Ou et~al\mbox{.}}{2018}]%
        {Ou2018}
\bibfield{author}{\bibinfo{person}{Weihua Ou}, \bibinfo{person}{Di Yuuan},
  {and} \bibinfo{person}{Yongfeng Cao}.} \bibinfo{year}{2018}\natexlab{}.
\newblock \showarticletitle{Object tracking based on online representative
  sample selection via noon-negative least square}.
\newblock \bibinfo{journal}{\emph{Multimed Tools Appl}} \bibinfo{number}{77}
  (\bibinfo{year}{2018}), \bibinfo{pages}{10569--10587}.
\newblock


\bibitem[\protect\citeauthoryear{Pavlichenko, Stelmakh, and
  Ustalov}{Pavlichenko et~al\mbox{.}}{2021}]%
        {Survey_45}
\bibfield{author}{\bibinfo{person}{Nikita Pavlichenko}, \bibinfo{person}{Ivan
  Stelmakh}, {and} \bibinfo{person}{Dmitry Ustalov}.}
  \bibinfo{year}{2021}\natexlab{}.
\newblock \showarticletitle{CrowdSpeech and VoxDIY: Benchmark Datasets for
  Crowdsourced Audio Transcription}.
\newblock \bibinfo{journal}{\emph{arXiv preprint arXiv:2107.01091}}
  (\bibinfo{year}{2021}).
\newblock


\bibitem[\protect\citeauthoryear{Peacock}{Peacock}{1983}]%
        {peacock1983two}
\bibfield{author}{\bibinfo{person}{John~A Peacock}.}
  \bibinfo{year}{1983}\natexlab{}.
\newblock \showarticletitle{Two-dimensional goodness-of-fit testing in
  astronomy}.
\newblock \bibinfo{journal}{\emph{Monthly Notices of the Royal Astronomical
  Society}} \bibinfo{volume}{202}, \bibinfo{number}{3} (\bibinfo{year}{1983}),
  \bibinfo{pages}{615--627}.
\newblock


\bibitem[\protect\citeauthoryear{Petersen, Minkkinen, and Esbensen}{Petersen
  et~al\mbox{.}}{2005}]%
        {Petersen2005}
\bibfield{author}{\bibinfo{person}{Lars Petersen}, \bibinfo{person}{Pentti
  Minkkinen}, {and} \bibinfo{person}{Kim~H. Esbensen}.}
  \bibinfo{year}{2005}\natexlab{}.
\newblock \showarticletitle{Representative sampling for reliability data
  analysis: Theory of Sampling}.
\newblock \bibinfo{journal}{\emph{Chemometrics and Intelligent Laboratory
  Systems}} \bibinfo{number}{77} (\bibinfo{year}{2005}),
  \bibinfo{pages}{261--277}.
\newblock


\bibitem[\protect\citeauthoryear{Phillips, Jiang, Narvekar, Ayyad, and
  O'Toole}{Phillips et~al\mbox{.}}{2011}]%
        {phillips2011otherrace}
\bibfield{author}{\bibinfo{person}{P~Jonathon Phillips}, \bibinfo{person}{Fang
  Jiang}, \bibinfo{person}{Abhijit Narvekar}, \bibinfo{person}{Julianne Ayyad},
  {and} \bibinfo{person}{Alice~J O'Toole}.} \bibinfo{year}{2011}\natexlab{}.
\newblock \showarticletitle{An other-race effect for face recognition
  algorithms}.
\newblock \bibinfo{journal}{\emph{ACM Transactions on Applied Perception
  (TAP)}} \bibinfo{volume}{8}, \bibinfo{number}{2} (\bibinfo{year}{2011}),
  \bibinfo{pages}{1--11}.
\newblock


\bibitem[\protect\citeauthoryear{Rahman, Balakrishnan, Murthy, Kutlu, and
  Lease}{Rahman et~al\mbox{.}}{2021}]%
        {Survey_90}
\bibfield{author}{\bibinfo{person}{Md~Mustafizur Rahman},
  \bibinfo{person}{Dinesh Balakrishnan}, \bibinfo{person}{Dhiraj Murthy},
  \bibinfo{person}{Mucahid Kutlu}, {and} \bibinfo{person}{Matthew Lease}.}
  \bibinfo{year}{2021}\natexlab{}.
\newblock \showarticletitle{An Information Retrieval Approach to Building
  Datasets for Hate Speech Detection}.
\newblock \bibinfo{journal}{\emph{arXiv preprint arXiv:2106.09775}}
  (\bibinfo{year}{2021}).
\newblock


\bibitem[\protect\citeauthoryear{Raji and Buolamwini}{Raji and
  Buolamwini}{2019}]%
        {Raji}
\bibfield{author}{\bibinfo{person}{I. Raji} {and} \bibinfo{person}{J.
  Buolamwini}.} \bibinfo{year}{2019}\natexlab{}.
\newblock \showarticletitle{Actionable auditing: investigating the impact of
  publicly naming biased performance results of commercial AI products}.
\newblock \bibinfo{journal}{\emph{AAAI/ACM Conf. AI, Ethics, and Society}}
  (\bibinfo{year}{2019}).
\newblock


\bibitem[\protect\citeauthoryear{Roberts}{Roberts}{1971}]%
        {Roberts1971}
\bibfield{author}{\bibinfo{person}{Harry~V. Roberts}.}
  \bibinfo{year}{1971}\natexlab{}.
\newblock \showarticletitle{Committee Selection by Statistical Sampling}.
\newblock \bibinfo{journal}{\emph{The American Statistician}}
  \bibinfo{number}{25} (\bibinfo{date}{Feb} \bibinfo{year}{1971}),
  \bibinfo{pages}{18--20}.
\newblock
Issue 1.


\bibitem[\protect\citeauthoryear{Roberts, Ramapuram, Ranjan, Kumar, Bautista,
  Paczan, Webb, and Susskind}{Roberts et~al\mbox{.}}{2021}]%
        {Survey_ICCV_1}
\bibfield{author}{\bibinfo{person}{Mike Roberts}, \bibinfo{person}{Jason
  Ramapuram}, \bibinfo{person}{Anurag Ranjan}, \bibinfo{person}{Atulit Kumar},
  \bibinfo{person}{Miguel~Angel Bautista}, \bibinfo{person}{Nathan Paczan},
  \bibinfo{person}{Russ Webb}, {and} \bibinfo{person}{Joshua~M Susskind}.}
  \bibinfo{year}{2021}\natexlab{}.
\newblock \showarticletitle{Hypersim: A photorealistic synthetic dataset for
  holistic indoor scene understanding}. In
  \bibinfo{booktitle}{\emph{Proceedings of the IEEE/CVF International
  Conference on Computer Vision}}. \bibinfo{pages}{10912--10922}.
\newblock


\bibitem[\protect\citeauthoryear{Romano, Bates, and Candes}{Romano
  et~al\mbox{.}}{2020}]%
        {romano2020achieving}
\bibfield{author}{\bibinfo{person}{Yaniv Romano}, \bibinfo{person}{Stephen
  Bates}, {and} \bibinfo{person}{Emmanuel Candes}.}
  \bibinfo{year}{2020}\natexlab{}.
\newblock \showarticletitle{Achieving equalized odds by resampling sensitive
  attributes}.
\newblock \bibinfo{journal}{\emph{Advances in Neural Information Processing
  Systems}}  \bibinfo{volume}{33} (\bibinfo{year}{2020}),
  \bibinfo{pages}{361--371}.
\newblock


\bibitem[\protect\citeauthoryear{Ros and Guillaume}{Ros and Guillaume}{2016}]%
        {ros2016dendis}
\bibfield{author}{\bibinfo{person}{Fr{\'e}d{\'e}ric Ros} {and}
  \bibinfo{person}{Serge Guillaume}.} \bibinfo{year}{2016}\natexlab{}.
\newblock \showarticletitle{DENDIS: A new density-based sampling for clustering
  algorithm}.
\newblock \bibinfo{journal}{\emph{Expert Systems with Applications}}
  \bibinfo{volume}{56} (\bibinfo{year}{2016}), \bibinfo{pages}{349--359}.
\newblock


\bibitem[\protect\citeauthoryear{Rosenbaum}{Rosenbaum}{2010}]%
        {Rosenbaum2010}
\bibfield{author}{\bibinfo{person}{Paul~R. Rosenbaum}.}
  \bibinfo{year}{2010}\natexlab{}.
\newblock \bibinfo{booktitle}{\emph{Design of Observational Studies}}.
\newblock \bibinfo{publisher}{Springer}.
\newblock
\showISBNx{ISBN: 978-1-4419-1212-1}


\bibitem[\protect\citeauthoryear{Ruf and Detyniecki}{Ruf and
  Detyniecki}{2021}]%
        {ruf2021implementing}
\bibfield{author}{\bibinfo{person}{Boris Ruf} {and} \bibinfo{person}{Marcin
  Detyniecki}.} \bibinfo{year}{2021}\natexlab{}.
\newblock \showarticletitle{Implementing Fair Regression In The Real World}.
\newblock \bibinfo{journal}{\emph{arXiv preprint arXiv:2104.04353}}
  (\bibinfo{year}{2021}).
\newblock


\bibitem[\protect\citeauthoryear{Ruths and Pfeffer}{Ruths and Pfeffer}{2014}]%
        {Ruths2014}
\bibfield{author}{\bibinfo{person}{Derek Ruths} {and} \bibinfo{person}{Jürgen
  Pfeffer}.} \bibinfo{year}{2014}\natexlab{}.
\newblock \showarticletitle{Social media for large studies of behavior}.
\newblock \bibinfo{journal}{\emph{Science}} \bibinfo{number}{346}
  (\bibinfo{year}{2014}), \bibinfo{pages}{1063--1064}.
\newblock
Issue 6213.


\bibitem[\protect\citeauthoryear{Shahbazi, Lin, Asudeh, and Jagadish}{Shahbazi
  et~al\mbox{.}}{2022}]%
        {shahbazi2022survey}
\bibfield{author}{\bibinfo{person}{Nima Shahbazi}, \bibinfo{person}{Yin Lin},
  \bibinfo{person}{Abolfazl Asudeh}, {and} \bibinfo{person}{HV Jagadish}.}
  \bibinfo{year}{2022}\natexlab{}.
\newblock \showarticletitle{A Survey on Techniques for Identifying and
  Resolving Representation Bias in Data}.
\newblock \bibinfo{journal}{\emph{arXiv preprint arXiv:2203.11852}}
  (\bibinfo{year}{2022}).
\newblock


\bibitem[\protect\citeauthoryear{Shankar, Halpern, Breck, Atwood, Wilson, and
  Sculley}{Shankar et~al\mbox{.}}{2017}]%
        {shankar2017geodiversity}
\bibfield{author}{\bibinfo{person}{Shreya Shankar}, \bibinfo{person}{Yoni
  Halpern}, \bibinfo{person}{Eric Breck}, \bibinfo{person}{James Atwood},
  \bibinfo{person}{Jimbo Wilson}, {and} \bibinfo{person}{D Sculley}.}
  \bibinfo{year}{2017}\natexlab{}.
\newblock \showarticletitle{No classification without representation: Assessing
  geodiversity issues in open data sets for the developing world}.
\newblock \bibinfo{journal}{\emph{arXiv preprint arXiv:1711.08536}}
  (\bibinfo{year}{2017}).
\newblock


\bibitem[\protect\citeauthoryear{Shaw}{Shaw}{2006}]%
        {shaw2006}
\bibfield{author}{\bibinfo{person}{William~T. Shaw}.}
  \bibinfo{year}{2006}\natexlab{}.
\newblock \showarticletitle{Sampling Student's T distribution-use of the
  inverse cumulative distribution function}.
\newblock \bibinfo{journal}{\emph{Journal of Computational Finance}}
  \bibinfo{number}{9} (\bibinfo{year}{2006}), \bibinfo{pages}{37}.
\newblock
Issue 4.


\bibitem[\protect\citeauthoryear{Simpson}{Simpson}{1949}]%
        {simpson1949measurement}
\bibfield{author}{\bibinfo{person}{Edward~H Simpson}.}
  \bibinfo{year}{1949}\natexlab{}.
\newblock \showarticletitle{Measurement of diversity}.
\newblock \bibinfo{journal}{\emph{nature}} \bibinfo{volume}{163},
  \bibinfo{number}{4148} (\bibinfo{year}{1949}), \bibinfo{pages}{688--688}.
\newblock


\bibitem[\protect\citeauthoryear{StatisticsHowto.Com}{StatisticsHowto.Com}{2022}]%
        {StatisticsHowto}
\bibfield{author}{\bibinfo{person}{StatisticsHowto.Com}.}
  \bibinfo{year}{2022}\natexlab{}.
\newblock \bibinfo{booktitle}{\emph{Representative Sample: Simple Definition,
  Examples}}.
\newblock
\urldef\tempurl%
\url{https://www.statisticshowto.com/representative-sample/}
\showURL{%
\tempurl}


\bibitem[\protect\citeauthoryear{Suresh and Guttag}{Suresh and Guttag}{2019a}]%
        {suresh2019framework}
\bibfield{author}{\bibinfo{person}{Harini Suresh} {and} \bibinfo{person}{John~V
  Guttag}.} \bibinfo{year}{2019}\natexlab{a}.
\newblock \showarticletitle{A framework for understanding unintended
  consequences of machine learning}.
\newblock \bibinfo{journal}{\emph{arXiv preprint arXiv:1901.10002}}
  \bibinfo{volume}{2} (\bibinfo{year}{2019}), \bibinfo{pages}{8}.
\newblock


\bibitem[\protect\citeauthoryear{Suresh and Guttag}{Suresh and Guttag}{2019b}]%
        {Suresh2019}
\bibfield{author}{\bibinfo{person}{Harini Suresh} {and} \bibinfo{person}{John~V
  Guttag}.} \bibinfo{year}{2019}\natexlab{b}.
\newblock \showarticletitle{A Framework for Understanding Unintended
  Consequences of Machine Learning}.
\newblock \bibinfo{journal}{\emph{arXiv:1901.10002v1}} (\bibinfo{year}{2019}).
\newblock


\bibitem[\protect\citeauthoryear{Taleb}{Taleb}{2007}]%
        {Taleb2007}
\bibfield{author}{\bibinfo{person}{Nassim~Nicholas Taleb}.}
  \bibinfo{year}{2007}\natexlab{}.
\newblock \bibinfo{booktitle}{\emph{The Black Swan: The Impact of the Highly
  Improbable}}.
\newblock \bibinfo{publisher}{Random House}.
\newblock
\showISBNx{ISBN 978-1400063512}


\bibitem[\protect\citeauthoryear{Taleb}{Taleb}{2020}]%
        {Taleb2020}
\bibfield{author}{\bibinfo{person}{Nassim~Nicholas Taleb}.}
  \bibinfo{year}{2020}\natexlab{}.
\newblock \bibinfo{booktitle}{\emph{Statistical Consequences of Fat Tails}}.
\newblock \bibinfo{publisher}{STEM Academic Press}.
\newblock
\showISBNx{ISBN 978-1-5445-0805-4}


\bibitem[\protect\citeauthoryear{Torralba and Efros}{Torralba and
  Efros}{2011}]%
        {Torralba2011}
\bibfield{author}{\bibinfo{person}{Antonio Torralba} {and}
  \bibinfo{person}{Alexei~A. Efros}.} \bibinfo{year}{2011}\natexlab{}.
\newblock \showarticletitle{Unbiased look at dataset bias}.
\newblock \bibinfo{journal}{\emph{CVPR}} (\bibinfo{year}{2011}),
  \bibinfo{pages}{1521--1528}.
\newblock


\bibitem[\protect\citeauthoryear{Uy, Pham, Hua, Nguyen, and Yeung}{Uy
  et~al\mbox{.}}{2019}]%
        {Uy2019}
\bibfield{author}{\bibinfo{person}{Mikaela~Angelina Uy},
  \bibinfo{person}{Quang-Hieu Pham}, \bibinfo{person}{Binh-Son Hua},
  \bibinfo{person}{Thanh Nguyen}, {and} \bibinfo{person}{Sai-Kit Yeung}.}
  \bibinfo{year}{2019}\natexlab{}.
\newblock \showarticletitle{Revisiting Point Cloud Classification: A New
  Benchmark Dataset and Classification Model on Real-World Data}.
\newblock \bibinfo{journal}{\emph{Proceedings of: IEEE/CVF International
  Conference on Computer Vision (ICCV)}} (\bibinfo{year}{2019}),
  \bibinfo{pages}{1588--1597}.
\newblock


\bibitem[\protect\citeauthoryear{Vaserstein}{Vaserstein}{1969}]%
        {vaserstein1969markovEMD1}
\bibfield{author}{\bibinfo{person}{Leonid~Nisonovich Vaserstein}.}
  \bibinfo{year}{1969}\natexlab{}.
\newblock \showarticletitle{Markov processes over denumerable products of
  spaces, describing large systems of automata}.
\newblock \bibinfo{journal}{\emph{Problemy Peredachi Informatsii}}
  \bibinfo{volume}{5}, \bibinfo{number}{3} (\bibinfo{year}{1969}),
  \bibinfo{pages}{64--72}.
\newblock


\bibitem[\protect\citeauthoryear{Yang, Qinami, Fei-Fei, Deng, and
  Russakovsky}{Yang et~al\mbox{.}}{2020}]%
        {Yang2020}
\bibfield{author}{\bibinfo{person}{Kaiyu Yang}, \bibinfo{person}{Klint Qinami},
  \bibinfo{person}{Li Fei-Fei}, \bibinfo{person}{Jia Deng}, {and}
  \bibinfo{person}{Olga Russakovsky}.} \bibinfo{year}{2020}\natexlab{}.
\newblock \showarticletitle{Towards Fairer Datasets: Filtering and Balancing
  the Distribution of the People Subtree in the ImageNet Hierarchy}.
\newblock \bibinfo{journal}{\emph{FAT* '20, january 27-30}}
  (\bibinfo{year}{2020}).
\newblock


\bibitem[\protect\citeauthoryear{Yu, Xu, Zhang, Zhao, Guan, and Tao}{Yu
  et~al\mbox{.}}{2021}]%
        {Survey_61}
\bibfield{author}{\bibinfo{person}{Hang Yu}, \bibinfo{person}{Yufei Xu},
  \bibinfo{person}{Jing Zhang}, \bibinfo{person}{Wei Zhao},
  \bibinfo{person}{Ziyu Guan}, {and} \bibinfo{person}{Dacheng Tao}.}
  \bibinfo{year}{2021}\natexlab{}.
\newblock \showarticletitle{Ap-10k: A benchmark for animal pose estimation in
  the wild}.
\newblock \bibinfo{journal}{\emph{arXiv preprint arXiv:2108.12617}}
  (\bibinfo{year}{2021}).
\newblock


\bibitem[\protect\citeauthoryear{Yuan, Ippolito, Nikolaev, Callison-Burch,
  Coenen, and Gehrmann}{Yuan et~al\mbox{.}}{2021}]%
        {Survey_64}
\bibfield{author}{\bibinfo{person}{Ann Yuan}, \bibinfo{person}{Daphne
  Ippolito}, \bibinfo{person}{Vitaly Nikolaev}, \bibinfo{person}{Chris
  Callison-Burch}, \bibinfo{person}{Andy Coenen}, {and}
  \bibinfo{person}{Sebastian Gehrmann}.} \bibinfo{year}{2021}\natexlab{}.
\newblock \showarticletitle{SynthBio: A Case Study in Human-AI Collaborative
  Curation of Text Datasets}.
\newblock \bibinfo{journal}{\emph{arXiv preprint arXiv:2111.06467}}
  (\bibinfo{year}{2021}).
\newblock


\bibitem[\protect\citeauthoryear{Z.~Obermeyer and Mullainan}{Z.~Obermeyer and
  Mullainan}{[n.\,d.]}]%
        {Obermeyer}
\bibfield{author}{\bibinfo{person}{C.~Vogeli Z.~Obermeyer, B.~Powers} {and}
  \bibinfo{person}{S. Mullainan}.} \bibinfo{year}{[n.\,d.]}\natexlab{}.
\newblock \showarticletitle{Dissecting racial bias in an algorithm used to
  manage the health of populations}.
\newblock \bibinfo{journal}{\emph{Science}} (\bibinfo{year}{[n.\,d.]}).
\newblock


\bibitem[\protect\citeauthoryear{Zhou, Kuscsik, Liu, Medo, Wakeling, and
  Zhang}{Zhou et~al\mbox{.}}{2010}]%
        {zhou2010solving}
\bibfield{author}{\bibinfo{person}{Tao Zhou}, \bibinfo{person}{Zolt{\'a}n
  Kuscsik}, \bibinfo{person}{Jian-Guo Liu}, \bibinfo{person}{Mat{\'u}{\v{s}}
  Medo}, \bibinfo{person}{Joseph~Rushton Wakeling}, {and}
  \bibinfo{person}{Yi-Cheng Zhang}.} \bibinfo{year}{2010}\natexlab{}.
\newblock \showarticletitle{Solving the apparent diversity-accuracy dilemma of
  recommender systems}.
\newblock \bibinfo{journal}{\emph{Proceedings of the National Academy of
  Sciences}} \bibinfo{volume}{107}, \bibinfo{number}{10}
  (\bibinfo{year}{2010}), \bibinfo{pages}{4511--4515}.
\newblock


\end{thebibliography}

\appendix
\newpage
\section{Additional Results for California} \label{Append:CA}
We show 5 fold cross validation (CV) results from applying models trained on California data to all other states. We show regression results from linear regression models using the continuous income target and classification results for logistic regression models using the binarized income (50,000 USD threshold). We compare miniature samples generated with stratified random sampling to coverage samples produced through density sampling. 

\begin{figure}[h!]
    \centering
    \begin{subfigure}[b]{0.99\textwidth}
        \includegraphics[width=1\textwidth]{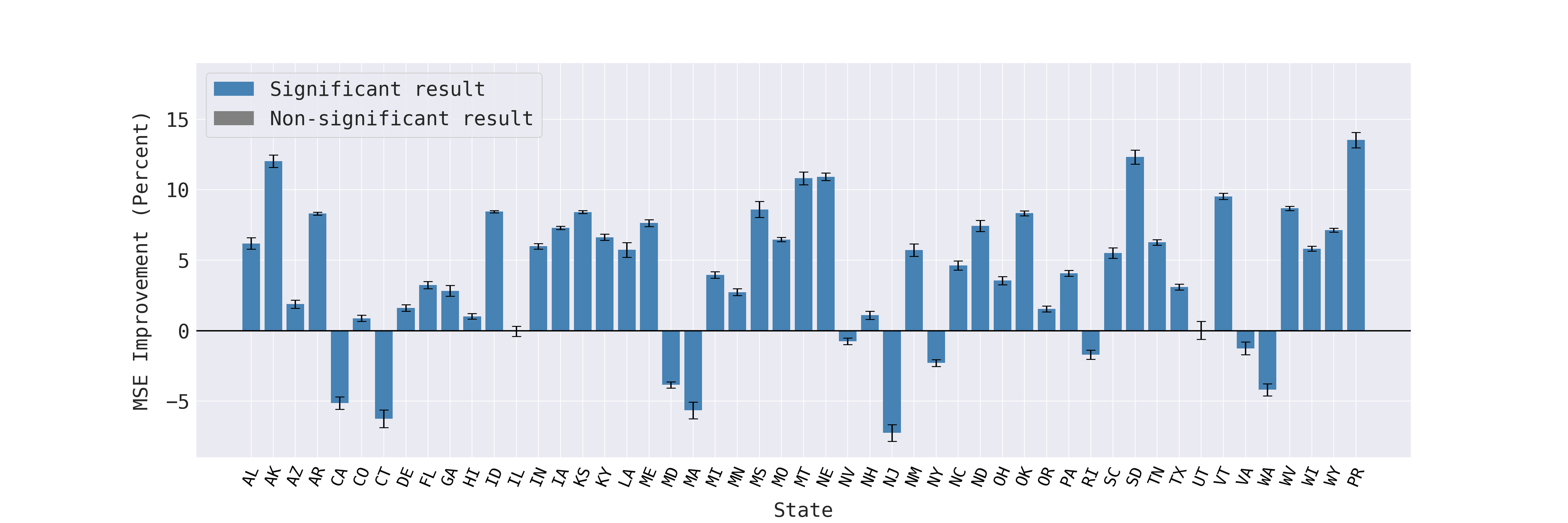}
        \caption{California regression.}
        \label{fig:}
    \end{subfigure}
    \vspace{\baselineskip}
    \begin{subfigure}[b]{0.99\textwidth}
        \includegraphics[width=1\textwidth]{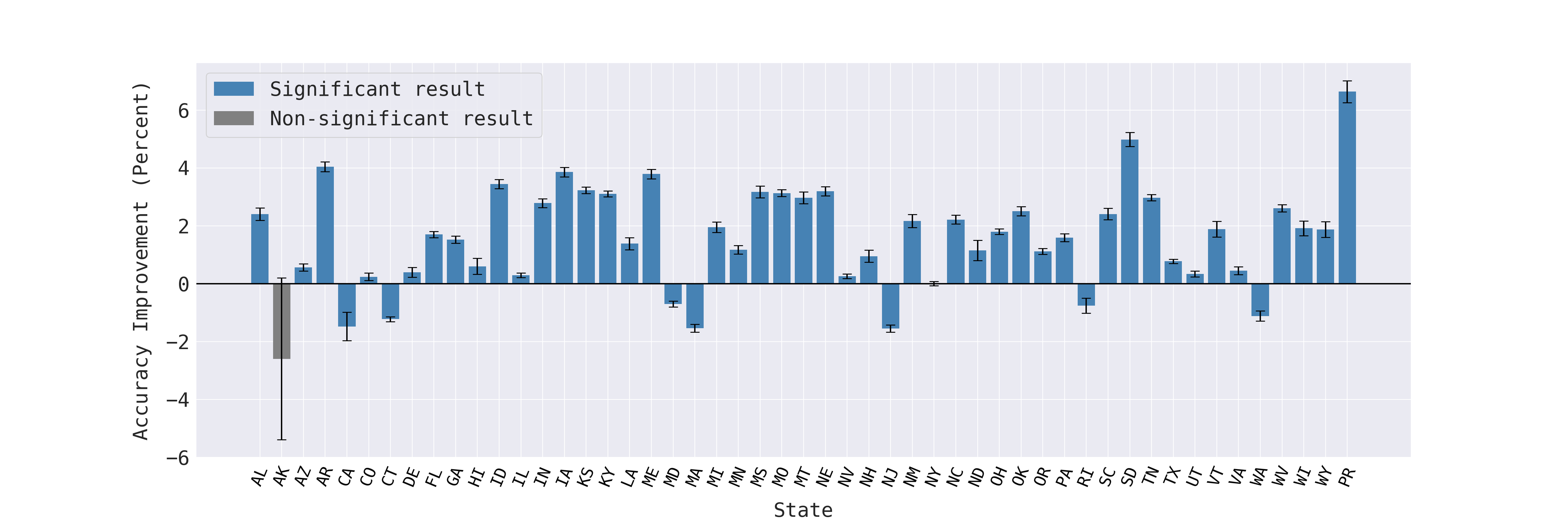}
        \caption{California classification.}
        \label{fig:}
    \end{subfigure}

    \caption{(a) Regression performance improvement on MSE when using coverage samples created with density sampling compared to using miniature samples created with proportional stratified sampling of the California training data. Error bars indicate the standard deviation. Positive values express that the MSE of the model trained on the coverage samples is improved over the MSE of the model trained on miniature samples. Negative values indicate that the coverage sampling deteriorates the performance compared to the miniature sampling. The mean MSE improvement across all states is 3.95$\%$. False Discovery Rate (FDR) adjusted paired t-tests have been performed on all interstate differences. Grey results (Illinois and Utah) indicate states where model differences are not significant to an $\alpha=0.05$ level. (b) Same as above for classification accuracy improvement on the binarized target value. The mean accuracy improvement across all states is 1.54$\%$. All performance differences except for Alaska and New York are significant to an $\alpha=0.05$ level when using a FDR adjusted paired sample t-test.}\label{fig:CA_Regression}
    \Description[Overview of performance on all US states for regression and classification models trained on coverage and miniature samples of the California data.]{The figure displays how the model trained on coverage samples performs better on the majority of the US states than the model trained on the miniature samples. This is especially the case for states dissimilar to California. On the other hand the model trained on miniature samples performs best on in-distribution data from California as well as several other states with demographic distributions similar to California.}
\end{figure}

Table \ref{tab:results_classification} shows in-distribution results for classification models trained using the binarized income with 50,000 USD threshold. Table \ref{tab:t-test_classification} shows the associated p-values while Table \ref{tab:t-test_regression} shows the p-values for the regression results in Table \ref{tab:results_regression}.

 \begin{table}[h!]
   \caption{Classification performance metrics evaluated with 5-fold cross validation of logistic regression models trained on the California data using the original binarized income (50,000 USD threshold) as target. The results are generated with the same procedure as in table \ref{tab:results_regression}. Accuracy is shown in addition to classification demographic disparity (CDD) and equal opportunity disparity (CEOD) for White / Native American individuals.}
  \label{tab:results_classification}
  \begin{tabular}{cccccccl}
    \toprule
    Training Data & Accuracy &  Parity (CDD) & Equality (CEOD) & Accuracy SD & Parity SD & Equality SD \\
    \midrule
    Full Census      & \textbf{0.7674} & 0.2510     &    0.2968     &   0.0027 & 0.0184 & 0.0700\\
    Miniature Sample &  0.7673         & 0.2666     &    0.3160     &   0.0028 & 0.0290 & 0.0793 \\
    Density Sample   &   0.7560 & \textbf{0.2096} & \textbf{0.2497}  &   0.0025 & 0.0182 & 0.0548\\
    DPP Sample       &  0.7634         & 0.2447     &    0.2938     &   0.0028 & 0.0220 & 0.0596 \\

  \bottomrule
\end{tabular}
\end{table}

 \begin{table}[h!]
 \caption{Paired sample t-test on classification model results shown in Table \ref{tab:results_classification}. P-values are adjusted for multiple testing by controlling the FDR using the  Benjamini-Hochberg procedure \cite{benjamini1995controlling}.}
  \label{tab:t-test_classification}
  \begin{tabular}{ccccl}
    \toprule
    Sample Comparison & Accuracy p-value & CDD p-value & CEOD p-value \\
    \midrule
    Full Census vs. Miniature & 0.7200  & 0.3198 & 0.3715  \\
    Density vs. Miniature     & 0.0007  & 0.0096 & 0.0249   \\
    DPP vs. Miniature         & 0.0023  & 0.2622 & 0.3605   \\
    DPP vs. Density           & 0.0031  & 0.0249 & 0.0755   \\
  \bottomrule
\end{tabular}
\end{table}

 \begin{table}[h!]
 \caption{Paired sample t-test on regression model results shown in Table \ref{tab:results_regression}. P-values are adjusted for multiple testing by controlling the False Discovery Rate (FDR) using the  Benjamini-Hochberg procedure \cite{benjamini1995controlling}.}
  \label{tab:t-test_regression}
  \begin{tabular}{cccl}
    \toprule
    Sample Comparison & MSE p-value & RDD p-value & REOD p-value \\
    \midrule
    Full Census vs. Miniature  & 0.0624  & 0.2962 & 0.1970    \\
    Density vs. Miniature      & 0.0002  & 0.0010 & 0.0102  \\
    DPP vs. Miniature          & 0.0192  & 0.0617 & 0.0102  \\
    DPP vs. Density            & 0.0010  & 0.0044 & 0.0332  \\
  \bottomrule
\end{tabular}
\end{table}

\section{Results for Massachusetts} \label{Append:MA}
We show 5 fold CV results from models trained on the Massachusetts data (n=40,114) applied to all other states. The Massachusetts data is significantly different from the California data both in terms of overall data size (n=40,114 vs. n=195,665), proportion of individuals with Bachelor's degree (49.1$\%$ vs. 38.7$\%$) and in terms of proportion of White individuals (82.3$\%$ vs. 61.8$\%$), but is similar on several other parameters like average weekly hours worked (37.4 vs. 37.9), proportion of government workers (12.5$\%$ vs. 14.9$\%$), proportion of individuals aged 10-33 (33.3$\%$ vs. 32.4$\%$) and proportion of married individuals (51.6$\%$ vs. 52.4$\%$). Results compare performances for models trained on miniature samples of the training data to models trained on coverage samples created with the density sampling approach. Both sample types have size 20$\%$ of the training data. We find similar results as with the California data, where for both regression and classification using coverage samples over miniature samples deteriorates performance in terms of MSE and accuracy on in-distribution states similar to the Massachusetts training state (most notably on states CA, CT, MD, MA and NJ). However, model performance on states dissimilar to Massachusetts (out-of-distribution) is improved with an average performance increase across all states (including the training state) of 3.27$\%$ for regression and 1.54$\%$ for classification. 

\begin{figure}[h!]
    \centering
    \begin{subfigure}[b]{0.99\textwidth}
        \includegraphics[width=1\textwidth]{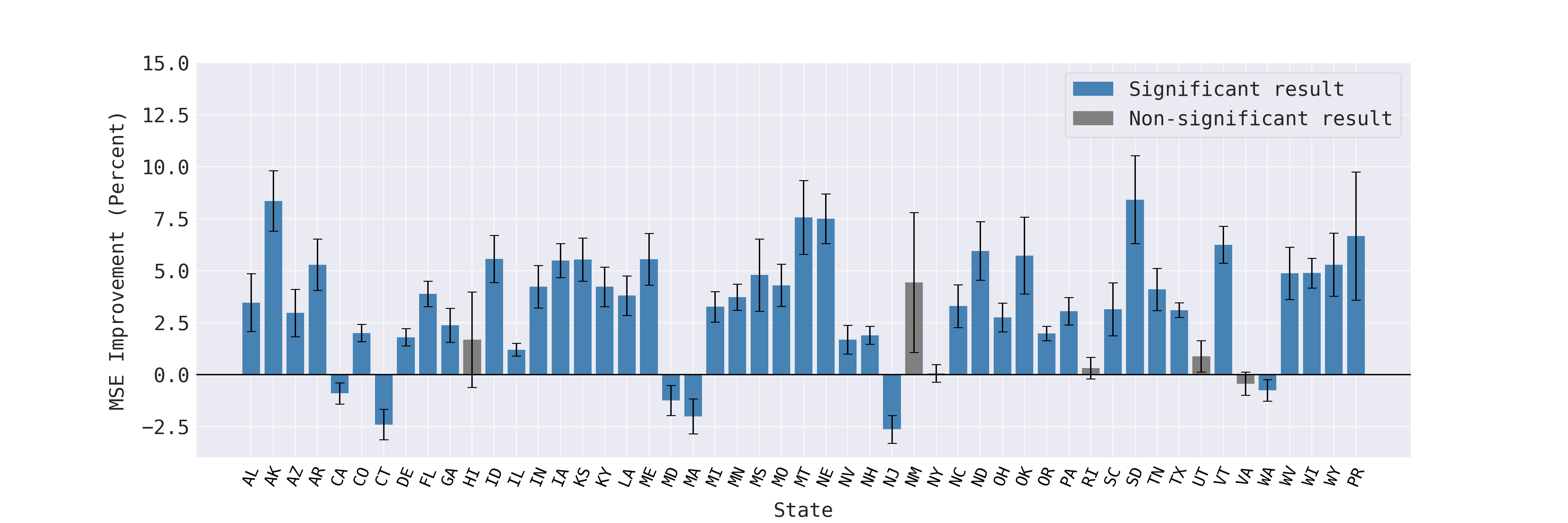}
        \caption{MA regression.}
        \label{fig:}
    \end{subfigure}
    \vspace{\baselineskip}
    \begin{subfigure}[b]{0.99\textwidth}
        \includegraphics[width=1\textwidth]{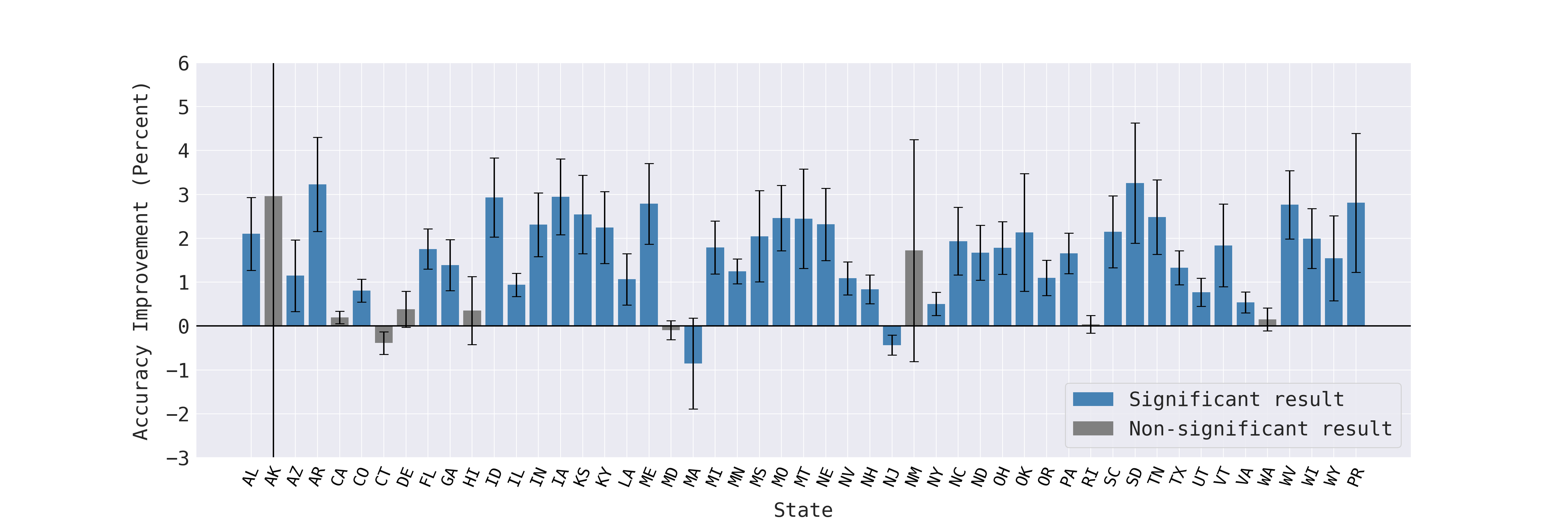}
        \caption{MA classification.}
        \label{fig:}
    \end{subfigure}

    \caption{(a) Regression performance improvement on MSE when using coverage samples compared to miniature samples of the Massachusetts training data. Mean MSE performance improvement across all states is 3.27$\%$. b) Same as above for classification. The mean accuracy improvement across all states is 1.54$\%$. FDR adjusted paired t-tests have been performed on all interstate differences. Grey results indicate states where model differences are not significant to an $\alpha=0.05$ level.}\label{fig:MA_Regression}
    \Description[Overview of performance on all US states for regression and classification models trained on coverage and miniature samples of the Massachusetts data.]{The figure displays how the model trained on coverage samples performs better on the majority of the US states than the model trained on the miniature samples. This is especially the case for states dissimilar to Massachusetts. On the other hand the model trained on miniature samples performs best on in-distribution data from Massachusetts as well as several other states with demographic distributions similar to Massachusetts.}
\end{figure}

\end{document}